\documentclass[twoside]{article}

\usepackage[preprint]{aistats2027}

\usepackage[round]{natbib}

\usepackage{graphicx}
\usepackage{float}
\usepackage{amsmath,amssymb,amsthm,mathtools}
\usepackage{dsfont}
\mathtoolsset{showonlyrefs}

\newtheorem{proposition}{Proposition}
\newtheorem{corollary}{Corollary}
\theoremstyle{definition}
\newtheorem{definition}{Definition}

\usepackage{pgfplots,pgfplotstable}
\usepgfplotslibrary{groupplots,fillbetween}
\usetikzlibrary{patterns}
\pgfplotsset{compat=1.18}

\usepackage{booktabs}
\usepackage{enumitem}

\usepackage{xcolor}

\definecolor{refblue}{HTML}{2E6DA4}

\usepackage[breaklinks]{hyperref}
\hypersetup{
    pdftitle={Designing Experiments for External Decision Makers},
    pdfauthor={Sahel Iqbal and Tom Rainforth},
    colorlinks=true,
    linkcolor=refblue,
    citecolor=refblue,
    urlcolor=refblue
}

\makeatletter
\renewcommand\hyper@natlinkbreak[2]{#1}
\makeatother

\DeclarePairedDelimiter\norm{\lVert}{\rVert}
\DeclareMathOperator*{\argmin}{arg\,min\,}
\DeclareMathOperator*{\argmax}{arg\,max\,}
\newcommand{\dd}{\mathrm{d}}

\DeclarePairedDelimiterX{\infdivx}[2]{[}{]}{#1\;\delimsize\|\;#2}
\newcommand{\kld}{\mathrm{KL}\infdivx*}

\begin{document}

\runningtitle{Designing Experiments for External Decision Makers}
\runningauthor{Iqbal and Rainforth}

\twocolumn[
\aistatstitle{Designing Experiments for External Decision Makers}
\aistatsauthor{Sahel Iqbal \And Tom Rainforth}
\aistatsaddress{University of Oxford \And University of Oxford}
]

\begin{abstract}
    \looseness=-1
    Standard Bayesian experimental design (BED) methods assume the
    experimenter alone will act on the data, choosing the best
    action given their own beliefs and goals. In practice, data are often acted
    on by external parties whose procedures the experimenter cannot change and
    may not know, particularly when the data are released publicly. To extend
    BED to such settings, we introduce usage-aware experimental design, a
    framework in which an adaptive design policy is trained to minimise the
    usage risk: the experimenter's expected loss when one or more external
    rules act on the collected data. These rules may be non-Bayesian or
    black-box, and when several may apply, we either average over them or take
    the worst case. We show that, unlike in standard BED, more informative data
    can be harmful under such rules, so the usage risk cannot in general be
    expressed as an expected uncertainty reduction; our framework thus strictly
    generalises standard BED. We then use it to tackle a range of problems that
    are difficult to address with existing approaches.
\end{abstract}

\section{INTRODUCTION}
\label{sec:intro}

Which data to collect to aid in decision making is a question that arises
throughout science, engineering, policy, and many other domains, and has
motivated the formal study of the optimal design of experiments for over a
century~\citep{smith1918standard, fisher1935design}. Bayesian experimental
design~(BED) offers a principled framework to answer this question: it models
the experiment probabilistically and chooses designs that minimise expected loss
over the possible outcomes~\citep{lindley1956measure, chaloner1995bayesian}.

\looseness=-1
In the foundational decision-theoretic formulation of BED~\citep{wald1950statistical, savage1951statistical, raiffa1961applied,
lindley1972bayesian}, there are two decisions: first, which experiments to
perform, as chosen by our design policy, and second, after observing the data, which downstream action
to take~(e.g., whether to approve a drug after a trial).
Here our design policy must plan ahead and try to maximise our ability to choose effective downstream actions.
Much of the BED literature has
abstracted away the second decision and instead sought experiments that maximally
reduce uncertainty about some variables of interest $z$~\citep{ryan2016review,
foster2021deep, rainforth2024modern, huan2024optimal}. 
More recently, a growing
body of work has made the second decision explicit, designing experiments with
reference to the expected future loss of optimal downstream actions rather than generic uncertainties~\citep{filstroff2024targeted, huang2024amortized, huang2026loss,
rossa2026actionbed, go2026goal}.

\looseness=-1
Both lines of work assume that the experimenter chooses the downstream action.
For loss-driven approaches this is explicit; for uncertainty-based ones it is
implicit in the equivalence between the expected loss of the Bayes-optimal
action and generalised entropies~\citep{degroot1962uncertainty,
dawid1998coherent, grunwald2004game, bickfordsmith2025rethinking}. Either way,
experiments are evaluated as if the action will be Bayes-optimal under the
experimenter's own beliefs and loss, which gives almost all existing BED
objectives their defining property: more information never hurts in
expectation~\citep{blackwell1953equivalent}.

However, in many applications,
the data are instead acted on by external decision makers, who apply their own
rules mapping data to actions. Sometimes a single rule is known in advance, such
as when a regulator applies a prespecified test to a trial run by a
pharmaceutical company, or a trained and frozen prediction model receives newly
acquired measurements. In other cases, the rule is not known in advance, such as when
it is unclear which of several decision makers will act, or when data are
released for public use and analysed by many parties, each with their own
procedure.

How, then, should the experimenter choose experiments when they are not the one
who will act on the data? Existing work answers this only for narrowly specified
rules, such as the Bayes act of a second analyst~\citep{jackson1980adversary,
lindley1991evidence, etzioni1993optimal, wolfson1997subjective} or a
prespecified test~\citep{ohagan2005assurance, ondra2019optimized,
rossell2013sequential}, and then only for static designs or via bespoke derivations.

To address this problem in the general case, we introduce a framework called
\emph{usage-aware experimental design}, in which experiments are designed
to account for how their data will subsequently be used, allowing for uncertainty
about who will act and which decision rules they will apply. The experimenter still
has preferences over the actions taken, expressed as a loss
function~\citep{savage1972foundations}, so we judge each design by its
\emph{usage risk}: the experimenter's expected loss when the downstream decision rule,
rather than their own Bayes act, chooses the action. The rule need only map data
to actions, so it may be another analyst's Bayes act, a non-Bayesian procedure
such as a significance test, or a black box such as a frozen prediction model.
When the rule is uncertain, we minimise either the average usage risk over
the plausible rules or its worst case. In each case, an amortised design policy
can be learned by simulation, which needs only evaluations of the rule.

We prove that usage-aware design \emph{strictly generalises} standard BED.
We decompose the usage risk into the risk under the Bayes act plus an
\emph{incompatibility cost}, which bounds the price of ignoring the rule and
vanishes when the rule is the experimenter's own Bayes act, recovering standard
BED. Building on Blackwell's characterisation, we then show that, for
posterior-based rules, more information never hurts if and only if the rule's
posterior risk, its expected loss as a function of the posterior, is concave:
that is, if it is itself a measure of uncertainty in the sense of
\citet{degroot1962uncertainty}. Standard BED thus corresponds exactly to rules
whose posterior risk is an uncertainty. For other rules, evidence that is
informative about $z$ need not be evidence the rule can exploit, and can even
steer it towards worse decisions; preferring less informative data is then not
irrational, since the experimenter's beliefs and loss are unchanged and only the
actor differs, yet no uncertainty-based criterion can capture it.

Empirically, on clinical trials, consumer persuasion, image
acquisition, and ranking and selection, usage-aware designs often
substantially outperform designs that ignore the rule, and collect
qualitatively different data.
Code to reproduce our experiments is available at
    \url{https://codeberg.org/Sahel13/usage-aware-experimental-design}.

\section{BACKGROUND}
\label{sec:background}

\looseness=-1
Let $z \in \mathcal{Z}$ denote an unknown state of the world.  In the
decision-theoretic formulation of BED~\citep{raiffa1961applied,
lindley1972bayesian}, an experimenter gathers data to support a downstream
action \(a \in \mathcal{A}\), whose consequences under \(z\) are measured by a
loss \(\ell(a, z)\), and aims to minimise the expected value of this loss.  The
experimenter encodes their initial beliefs about $z$ in a prior $p(z)$, chooses
a design $x \in \mathcal{X}$, observes an outcome \(y \sim p(y \mid z, x)\), and
updates their beliefs to the posterior \(p(z \mid x, y)\).  They then take the
action minimising the posterior expected loss, known as the Bayes-optimal action
or simply the \emph{Bayes act}. Since the outcome is unknown when the design is
chosen, a design is judged by the expected loss of the Bayes act under the
predictive distribution \(p(y \mid x)\):
\begin{equation}
    \label{eq:standard-bed-formulation}
    B_{\ell}(x) \coloneq \mathbb{E}_{p(y \mid x)} \left[ \inf_{a \in \mathcal{A}}
    \mathbb{E}_{p(z \mid x, y)}[\ell(a, z)] \right].
\end{equation}
This is the \emph{Bayes risk} of the design $x$, and minimising it is the
classical BED objective.

Much of BED instead minimises the \emph{expected posterior uncertainty}~(EPU), without reference to a downstream action~\citep{rainforth2024modern,huan2024optimal},
\begin{equation}
  \label{eq:epu}
  \mathrm{EPU}_{\mathbb{H}}(x) \coloneq \mathbb{E}_{p(y \mid x)} \Bigl[ \mathbb{H}\bigl[p(z \mid x, y)\bigr] \Bigr],
\end{equation}
where $ \mathbb{H}$ denotes some uncertainty measure. The two views are
equivalent whenever \(\mathbb{H}\) is a generalised
entropy~\citep{degroot1962uncertainty,dawid1998coherent,grunwald2004game},
defined for a distribution $q \in \mathcal{P}_{z}$ as
\begin{equation}
    \label{eq:generalized-entropy}
    \mathbb{H}\left[q\right] \coloneq \mathbb{E}_{Z\sim q}\left[s(q,Z)\right],
\end{equation}
\looseness=-1
where $s$ is a proper scoring rule on $\mathcal{P}_{z}$, the set of probability distributions on $\mathcal{Z}$.
The equivalence holds because the loss of the Bayes act under $q$, $a^{*}(q) \in \argmin_{a \in \mathcal{A}}
    \mathbb{E}_{Z\sim q}[\ell(a, Z)]$, assumed throughout to exist, is itself a proper scoring rule: for any $\ell$, setting $s(q,Z)=\ell(a^{*}(q),Z)$ gives
\begin{align}
    \mathbb{H}_{\ell} \left[q\right] = \inf_{a \in \mathcal{A}} \mathbb{E}_{Z\sim q}[\ell(a, Z)] = \mathbb{E}_{Z\sim q}\left[\ell(a^{*}(q),Z)\right].
\end{align}
The Bayes risk is thus the uncertainty, as measured by
\(\mathbb{H}_{\ell}\), that is expected to remain after the experiment. 
Following \citet{degroot1962uncertainty}, the defining property of a measure of
uncertainty is that no experiment can increase it in expectation. Every
generalised entropy has this property, and for finite $\mathcal{Z}$ the converse
also holds~\citep{grunwald2004game, bect2026uncertainty}. Specifying an
uncertainty measure directly, as is common in practice, therefore
\emph{implicitly} specifies a set of compatible downstream decision problems,
and assumes that the action eventually taken is the Bayes act for one of them.
When \(\mathbb{H}\) is the Shannon~(or differential)
entropy~\citep{shannon1948mathematical}, minimising the EPU is equivalent to
maximising the popular expected information
gain~\citep[EIG,][]{lindley1956measure}.

The Bayes risk extends naturally to sequential experiments, in which later
designs can depend on earlier observations. Over $T$ steps, let
$D_{t} \coloneq (x_{s}, y_{s})_{s=1}^{t} \in \mathcal{D}_{t}$ denote the collected data, or history, with
$D_{0} = \emptyset$. A \emph{design policy} $\pi_{d} \in \Pi_{d}$ selects each
$x_{t}$ given $D_{t-1}$, possibly at random, and induces a distribution
$p(D_{T} \mid z; \pi_{d})$ over histories for each state \(z\). The Bayes
risk is now a function of the design policy:
\begin{align}
    \label{eq:sequential-bayes-risk}
    B_{\ell}(\pi_{d}) &\coloneq \mathbb{E}_{p(D_{T}; \pi_{d})}
    \left[\mathbb{H}_{\ell}[p_{z}(\cdot \mid D_{T})]\right],
\end{align}
\looseness=-1
where $\mathbb{H}_{\ell}[p_{z}(\cdot \mid D_{T})]$ is the minimal achievable
expected loss under the posterior $p_{z}(\cdot \mid D_{T})$. Given $D_{T}$, the
posterior does not depend on the design policy, which thus affects $B_{\ell}$
only through the distribution of histories~\citep{phillips2026bayesian}.

\section{EXPERIMENTAL DESIGN FOR EXTERNAL DECISION MAKERS}
\label{sec:methodology}

The standard BED formulations in Section~\ref{sec:background} assume
that the experimenter both chooses the design and acts on the
outcome, in each case according to their own beliefs and loss.
\citet{kadane2014toward} accordingly describe classical BED as a ``single-party,
two-stage'' decision problem.

We now generalise this to settings in which the action is chosen by a fixed
rule rather than by the experimenter's Bayes act. Most often, the data are acted
on by someone else, possibly someone the experimenter cannot identify in
advance, as when a regulator assesses a trial or data are released publicly.
But the experimenter can also be bound to a rule when acting themselves, for
example by a legal obligation on how to act on the results of a safety
inspection, or by a frequentist analysis plan fixed before the data are seen.
Either way, the experimenter cannot replace the rule with their own Bayes act.

To formalise such settings, we introduce the notion of a
\emph{downstream decision
maker}~(DM) that is either external to the experimenter or represents the fixed rule the experimenter will later follow.
The experimenter still has
preferences over the actual actions taken, represented as in Section~\ref{sec:background}
by their own prior \(p(z)\) and loss \(\ell(a, z)\). What they no longer freely control
is the action itself. 
Instead, the DM will follow some \emph{action policy} \(\pi_{a} \in
\Pi_{a}\), a measurable rule mapping the completed history \(D_{T}\) to an
action in \(\mathcal{A}\). The loss incurred by \(\pi_{d}\), from the perspective of the \emph{experimenter}, is then
\(\ell(\pi_{a}(D_{T}), z)\),
in place of the loss of the Bayes act in~\eqref{eq:sequential-bayes-risk}.

Averaging this loss under the experimenter's beliefs gives the
\emph{usage risk} of \(\pi_{d}\) under \(\pi_{a}\),
\begin{equation}
    \label{eq:fixed-rule-risk}
    R_{\ell}(\pi_{d}, \pi_{a}) \coloneq \mathbb{E}_{p(z, D_{T}; \pi_{d})}
    \bigl[\ell(\pi_{a}(D_{T}), z)\bigr],
\end{equation}
where we assume throughout that the losses are integrable and the Bayes risk is
finite.\footnote{\looseness=-1 The loss
may also include a cost of the data collected, as in
Section~\ref{sec:destructive-lot-inspection}, by replacing $\ell(a, z)$ with
$\ell(a, z) + c(D_{T})$. This term enters the usage risk and the Bayes
risk identically, so the decomposition in Section~\ref{sec:incompatibility} is
unaffected, and Proposition~\ref{prop:blackwell-safe-rule} concerns the
informational part $\ell$.} 
For a fixed $\pi_{a}$, the experimenter's design problem is then to find
\begin{equation}
    \label{eq:fixed-rule-objective}
    \pi_{d}^{*} \in \argmin_{\pi_{d}\in\Pi_{d}}
    R_{\ell}(\pi_{d},\pi_{a}).
\end{equation}
Standard BED is the special case in which the DM applies the experimenter's
own Bayes act.
Indeed,~\citet{rossa2026actionbed} showed that, under mild assumptions, optimising the design policy jointly with an unrestricted action policy recovers the Bayes-risk minimiser:
\begin{equation}
    \label{eq:action-bed-formulation}
    \argmin_{\pi_{d} \in \Pi_{d}} \inf_{\pi_{a}} R_{\ell}(\pi_{d}, \pi_{a}) = \argmin_{\pi_{d} \in \Pi_{d}} B_{\ell}(\pi_{d}).
\end{equation}
By contrast,~\eqref{eq:fixed-rule-objective}
fixes \(\pi_{a}\) to the rule the DM will actually apply.
Each $\pi_{d}^{*}$ is still Bayes-optimal from the experimenter's perspective, given that this rule will be applied, as it is chosen under their own beliefs and loss.
But since $\pi_{a}$ need not be Bayesian, $\pi_{d}^{*}$ is \emph{conditionally} Bayes-optimal given the DM, rather than \emph{jointly} Bayes-optimal with the action.

\paragraph{Uncertainty over the downstream rule.}
The experimenter may also be uncertain about the DM's rule, such as
how the DM will model the data, which criterion they will optimise, or simply which procedure they will follow.
We parameterise these uncertain aspects by \(\psi \in \Psi\), writing the action policy as
\(\pi_{a}^{\psi}\), and extend the prior to $p(z,\psi)$.\footnote{$z$ and $\psi$ may be dependent, for example if the DM has access to data that the experimenter does not, which bears on both $z$ and the DM's choice of rule.}
The parameters may include discrete components, such as which of several
procedures is applied. We assume that \(\psi\) is unobserved during data
collection, so that \(p(z, \psi, D_{T}; \pi_{d}) = p(z, \psi)\, p(D_{T} \mid z; \pi_{d})\),
and that \((\psi, D_{T}) \mapsto \pi_{a}^{\psi}(D_{T})\) is jointly measurable. The
loss is now \(\ell(\pi_{a}^{\psi}(D_{T}), z)\), so the expected loss of
\(\pi_{d}\) is
\begin{equation}
    \label{eq:bayes-average-objective}
    \bar{R}_{\ell}(\pi_{d})
    \coloneq \mathbb{E}_{p(z, \psi, D_{T}; \pi_{d})}
    \bigl[\ell(\pi_{a}^{\psi}(D_{T}), z)\bigr],
\end{equation}
and the optimal design policy is again its minimiser.
This recovers~\eqref{eq:fixed-rule-objective} when \(\psi\) is known in advance.

The same objective covers data used by many DMs, when the experimenter wants to minimise the average loss of the DMs' actions.
Here \(p(\psi)\) is the proportion of DMs applying each rule, weighted, if desired, by how much the experimenter cares about each DM's actions.

\paragraph{Robustness to downstream-rule uncertainty.}
A prior \(p(\psi)\) may be difficult to justify, or the experimenter may want
protection against every plausible value of \(\psi\). In such cases, a more
suitable objective may be the largest usage risk over \(\Psi\),
\begin{equation}
    \label{eq:minimax-objective}
    R_{\ell}^{\mathrm{mm}}(\pi_{d}) \coloneq
    \sup_{\psi \in \Psi}
    \mathbb{E}_{p(z, D_{T} \mid \psi; \pi_{d})}
    \bigl[\ell(\pi_{a}^{\psi}(D_{T}), z)\bigr],
\end{equation}
which we minimise.
Unlike~\eqref{eq:bayes-average-objective}, this criterion is not an expected
loss, but it is a standard alternative for robust
decision making. It is also where randomised design policies
matter. Randomisation cannot
improve~\eqref{eq:fixed-rule-objective} or~\eqref{eq:bayes-average-objective},
which are linear in the distribution over histories, but it can strictly
reduce the supremum in~\eqref{eq:minimax-objective}~\citep[Ch. 5]{berger1985statistical}.

\paragraph{Policy learning.}
Each objective above is built from expectations of a loss over simulated
trajectories: the state and, where relevant, the rule parameters are drawn from
their priors, the design policy generates the history, and the loss is read off
by applying the rule to it. We therefore typically parameterise \(\pi_{d}\) by a neural
network and train it by stochastic gradient descent on Monte Carlo estimates of
the objective. The rule enters only through its evaluations on simulated
data, so we can use pathwise gradient estimators when the policy and simulator
are reparameterisable and the rule is differentiable, and score-function
estimators otherwise, including for black-box
rules~\citep{mohamed2020montecarlo}. Details are given in Appendix~\ref{app:policy-learning}.

\section{THEORETICAL INSIGHTS}
\label{sec:compatibility}

Fixing the downstream rule changes both the structure of the objective and the
value of information, and we now examine these in turn.

\subsection{Design-Dependent Incompatibility}
\label{sec:incompatibility}

\looseness=-1
From the experimenter's perspective, the natural benchmark for how well a rule
uses the data is the minimal posterior expected loss, $\mathbb{H}_{\ell}[p_{z}(\cdot \mid
D_{T})]$, which is the loss they would expect from the same data if they chose
the action themselves. The \emph{posterior regret} of an action policy $\pi_{a}$
is
\begin{equation}
    \Delta_{\pi_{a}}(D_{T})
    \coloneq
    \mathbb{E}_{p_{z}(\cdot \mid D_{T})}[\ell(\pi_{a}(D_{T}), z)]
    - \mathbb{H}_{\ell}[p_{z}(\cdot \mid D_{T})],
\end{equation}
which is non-negative, and zero for every $D_{T}$ when $\pi_{a}$ is a Bayes-optimal action policy, $\pi_{a}^{\mathrm{B}}$.
The usage risk can then be expressed as
\begin{align}
    \label{eq:deployment-risk-decomposition}
    R_{\ell}(\pi_{d}, \pi_{a}) & = B_{\ell}(\pi_{d}) + C_{\ell}(\pi_{d}, \pi_{a}), \\
    C_{\ell}(\pi_{d}, \pi_{a})
    & \coloneq
    \mathbb{E}_{p(D_{T}; \pi_{d})}[\Delta_{\pi_{a}}(D_{T})].
\end{align}
The \emph{incompatibility cost} $C_{\ell}$ is the expected excess loss the
experimenter incurs because the fixed rule, rather than the Bayes act, chooses
the action. It depends on the design policy, which controls the distribution of
datasets to which the rule is applied.

\(C_{\ell}\) takes familiar forms for some common losses. Under squared error, with
\(\mathcal{A} = \mathcal{Z} = \mathbb{R}^{d}\) and
\(\ell(a, z) = \norm{a - z}^{2}\), the Bayes act is the posterior mean
\(m_{T} = \mathbb{E}[z \mid D_{T}]\), and expanding the square gives
\(
    \Delta_{\pi_{a}}(D_{T}) = \norm{\pi_{a}(D_{T}) - m_{T}}^{2}.
\)
Hence \(C_{\ell}\) is the expected squared distance between the rule's estimate and
the posterior mean. Under log loss, the action is a distribution on \(\mathcal{Z}\)
and \(\ell(a, z) = -\log a(z)\). The Bayes act is the posterior itself,
\(\mathbb{H}_{\ell}[p_{z}(\cdot \mid D_{T})]\) is the Shannon entropy of the posterior, and
\(
    \Delta_{\pi_{a}}(D_{T}) = \kld{p_{z}(\cdot \mid D_{T})}{\pi_{a}(D_{T})},
\)
so that \(C_{\ell}\) is the expected KL divergence from the experimenter's posterior to
the distribution the rule reports. In both cases, \(C_{\ell}\) measures the expected
discrepancy between what the rule does with the data and what a Bayes-optimal
DM would do.

The incompatibility cost also bounds the price of designing as though the
downstream rule were Bayes-optimal.
\begin{proposition}[Cost of ignoring the downstream rule]
    \label{prop:cost-of-ignoring-rule}
    Let $\pi_{d}^{\mathrm{B}}\in\argmin_{\pi_{d}\in\Pi_{d}}B_{\ell}(\pi_{d})$ and
    $\pi_{d}^{*}\in\argmin_{\pi_{d}\in\Pi_{d}}R_{\ell}(\pi_{d},\pi_{a})$.
    If these minimisers exist,
    \begin{equation}
        \label{eq:rule-ignorance-bound}
        0
        \leq R_{\ell}(\pi_{d}^{\mathrm{B}},\pi_{a})
             - R_{\ell}(\pi_{d}^{*},\pi_{a})
        \leq C_{\ell}(\pi_{d}^{\mathrm{B}},\pi_{a}).
    \end{equation}
\end{proposition}
The first inequality holds because $\pi_{d}^{*}$ minimises the usage risk
under $\pi_{a}$. The second follows
from~\eqref{eq:deployment-risk-decomposition} applied to $\pi_{d}^{\mathrm{B}}$,
since $B_{\ell}(\pi_{d}^{\mathrm{B}}) \le B_{\ell}(\pi_{d}^{*}) \le
R_{\ell}(\pi_{d}^{*},\pi_{a})$ and $C_{\ell} \geq 0$. Thus, the
price of ignoring the rule is at most the incompatibility cost of the design
that ignores it. Conversely, a rule may have large regret but leave the
optimal design unchanged if $C_{\ell}(\pi_{d},\pi_{a})$ is constant across
designs.

\subsection{When Does More Information Help?}
\label{sec:value-of-information}

We now ask how the usage risk changes when the data become more
informative. We measure informativeness in the sense of
\citet{blackwell1951comparison, blackwell1953equivalent}, which does not depend
on a particular loss or prior.

\begin{definition}[Blackwell-more-informative]
    \label{def:blackwell}
    For $j\in\{0,1\}$, let $D_{T}^{j}$ denote the dataset generated by the policy
    $\pi_{d}^{j}$. We say that $\pi_{d}^{1}$ is \emph{Blackwell-more-informative} than
    $\pi_{d}^{0}$, denoted $\pi_{d}^{1}\succeq_{\mathrm{B}}\pi_{d}^{0}$, if there exists a
    Markov kernel $G(\dd D_{T}^{0}\mid D_{T}^{1})$, independent of $z$, such that
    \begin{equation}
        p(\dd D_{T}^{0}\mid z;\pi_{d}^{0})
        = \int G(\dd D_{T}^{0}\mid D_{T}^{1})
        \, p(\dd D_{T}^{1}\mid z;\pi_{d}^{1}),
    \end{equation}
    for every \(z \in \mathcal{Z}\).
    Thus, the dataset produced by $\pi_{d}^{0}$ can be simulated by stochastically
    post-processing the dataset produced by $\pi_{d}^{1}$, without access to the
    unknown state. Such post-processing is called a \emph{garbling}; informally, it adds noise to the original data.
\end{definition}

\looseness=-1
Blackwell's theorem~\citep{blackwell1953equivalent} shows that, for finite
$\mathcal{Z}$ and bounded losses, this is precisely the condition under which $\pi_{d}^{1}$ attains no larger Bayes risk
than $\pi_{d}^{0}$ in every decision problem, that is, for every prior and
every loss that does not depend on the data. In particular,
$\pi_{d}^{1}\succeq_{\mathrm{B}}\pi_{d}^{0}$ implies
$B_{\ell}(\pi_{d}^{1})\leq B_{\ell}(\pi_{d}^{0})$. The guarantee relies on the action being
reoptimised at every posterior. A non-Bayes-optimal rule does not do this, so
for such rules the theorem gives no assurance that more informative data will not hurt.

When, then, do more informative data help under a given rule, and when can they hurt?
We show that they can hurt if and only if the rule's posterior expected loss is not concave in the posterior.

Suppose that $\mathcal{Z}$ is finite, that $p(z)$ has full support, and that
$\pi_{a}(D_{T}) = f(b_{T})$ for some function
$f$ of the posterior $b_{T} \coloneq p_{z}(\cdot \mid D_{T}) \in
\mathcal{P}_{z}$. This covers any rule that acts on a statistic
that the posterior determines.
The usage risk is then the expectation over $D_{T}$
of the rule's \emph{posterior risk}
\begin{equation}
    r_{\pi_{a}}(b)
    \coloneq
    \mathbb{E}_{b(z)}[\ell(f(b), z)].
    \label{eq:rule-induced-posterior-risk}
\end{equation}

\begin{proposition}[Blackwell monotonicity for fixed rules]
    \label{prop:blackwell-safe-rule}
    Fix $f$ and let $\pi_{a} = f(b_{T})$. Then
    \begin{equation}
        R_{\ell}(\pi_{d}^{1},\pi_{a})\leq R_{\ell}(\pi_{d}^{0},\pi_{a})
    \end{equation}
    for all statistical experiments with
    $\pi_{d}^{1}\succeq_{\mathrm{B}}\pi_{d}^{0}$ if and only if
    $r_{\pi_{a}}$ is concave on $\mathcal{P}_{z}$. Here, statistical
    experiments are all Markov kernels from $\mathcal{Z}$ to a data space,
    for which we keep the notation $\pi_{d}$.
\end{proposition}

The proof, in Appendix~\ref{app:proofs}, adapts two classical results to
$r_{\pi_{a}}$: the convex-order characterisation of \citet{blackwell1953equivalent} for
sufficiency, and the splitting construction of \citet{kamenica2011bayesian}
for necessity.

\looseness=-1
If \(f\) returns the
Bayes act under \(b\), then $r_{\pi_{a}} = \mathbb{H}_{\ell}$, which is
concave as an infimum of functions that are affine in $b$, and Blackwell's theorem is
recovered. Other rules need not choose the best action at every belief, so
their posterior risk can fail to be concave, and more informative data can then push the
posterior towards beliefs where they err.
\citet{callahan2026misinformation} encounter the same failure when the
experimenter's own inference is misspecified, a special case
of~\eqref{eq:deployment-risk-decomposition}; where they replace monotonicity
with an axiom, Proposition~\ref{prop:blackwell-safe-rule} gives the exact
condition under which it holds.

\looseness=-1
A key consequence is that some problems expressible in our framework
\emph{cannot} be expressed as standard BED problems, under any loss.

\begin{corollary}[Usage risk is not a Bayes risk in general]
    \label{cor:epu-representation}
    Under the assumptions of Proposition~\ref{prop:blackwell-safe-rule}, let
    $\pi_{a}$ be an action policy for which $r_{\pi_{a}}$ is not concave. Then
    there exist experiments $\pi_{d}^{0}$ and $\pi_{d}^{1}$
    with $\pi_{d}^{1} \succeq_{\mathrm{B}} \pi_{d}^{0}$ such that, for every
    action space $\mathcal{A}'$ and loss
    $\ell' \colon \mathcal{A}' \times \mathcal{Z} \to (-\infty, +\infty]$, including
    $\ell$ itself,
    \begin{equation}
        R_{\ell}(\pi_{d}^{1},\pi_{a})
        >
        R_{\ell}(\pi_{d}^{0},\pi_{a})
        \text{ but }
        \, B_{\ell'}(\pi_{d}^{1})
        \leq
        B_{\ell'}(\pi_{d}^{0}),
    \end{equation}
    where $B_{\ell'}$ is the Bayes risk~\eqref{eq:sequential-bayes-risk} with
    $\ell'$ and $\mathcal{A}'$ in place of $\ell$ and $\mathcal{A}$.
    Consequently, no standard BED criterion of the form
    of~\eqref{eq:sequential-bayes-risk}, for any loss, induces the same
    ordering as the usage risk under $\ell$, over the class of all
    statistical experiments or over any class containing $\pi_{d}^{0}$ and
    $\pi_{d}^{1}$.
\end{corollary}
\looseness=-1
Its proof (Appendix~\ref{app:proofs}) uses only the concavity of
$\mathbb{H}_{\ell'}$, so it extends to the EPU under every measure of
uncertainty in the sense of \citet{degroot1962uncertainty}, as all are
concave~\citep{bect2026uncertainty}; this covers the EIG. Conversely, a concave
$r_{\pi_{a}}$ makes $R_{\ell} = \mathbb{E}[r_{\pi_{a}}(b_{T})]$ an EPU.
Usage-aware design thus strictly generalises BED, departing from it
exactly when the rule's posterior risk is not a measure of uncertainty.

\looseness=-1
The concavity condition also applies when the rule is uncertain, and in one
special case it has a more encouraging consequence. Suppose that $\psi$ is
independent of $z$, so that, by the factorisation in
Section~\ref{sec:methodology}, the data carry no information about the rule,
and that each $\pi_{a}^{\psi}$ acts on the
posterior $b_{T}$. Then $\bar{R}_{\ell}(\pi_{d}) = \mathbb{E}[\bar{r}(b_{T})]$,
where $\bar{r} \coloneq \mathbb{E}_{p(\psi)}[r_{\pi_{a}^{\psi}}]$ is the
mixture of the rules' posterior risks. Since the proof
of Proposition~\ref{prop:blackwell-safe-rule} uses only the identity
$R_{\ell}(\pi_{d}, \pi_{a}) = \mathbb{E}[r_{\pi_{a}}(b_{T})]$, which
$\bar{R}_{\ell}$ and $\bar{r}$ also satisfy, $\bar{R}_{\ell}$ is
Blackwell-monotone if and only if $\bar{r}$ is concave. Averaging over rules
smooths the posterior risk, so $\bar{r}$ can be concave even when no
$r_{\pi_{a}^{\psi}}$ is: when the data carry no information about which rule
will be applied, uncertainty about who will act on them can sometimes restore
the guarantee that more information does not hurt, at least on average.

\section{RELATED WORK}
\label{sec:related-work}

To our knowledge, designing experiments for a downstream rule that the
experimenter neither chooses nor necessarily knows has not been studied in
this generality. Some works,
however, study the special case in which the rule is the Bayes act of a second
Bayesian agent who differs from the experimenter in prior or
loss~\citep{jackson1980adversary, lindley1991evidence, etzioni1993optimal,
desantis2017decision, desantis2021predictive}. \citet{overstall2022bayesian}
consider the analogous case in which the second Bayesian fits a different model.
When several Bayesians may act on the data, \citet{lodh1993experimental} and
\citet{wolfson1997subjective} minimise a weighted average of their risks, which
gives static instances of~\eqref{eq:bayes-average-objective}. These works
derive a static design for a specified second Bayesian; our rules need not be
Bayes acts at all, and our policies are adaptive.

The problem of designing for a second Bayesian has also been studied
independently in economics as Bayesian persuasion: a sender commits to a
signal structure, and a receiver observes its realisation and takes the Bayes
act under their own prior and loss~\citep{kamenica2011bayesian}. Later work
admits receivers who hold a different worldview~\citep{galperti2019persuasion},
update by a non-Bayesian rule~\citep{declippel2022nonbayesian}, or apply a
frequentist test to an experiment drawn from a restricted feasible
set~\citep{ball2021experimental}. These works characterise a single, static
optimal signal analytically, for one known receiver and for models simple
enough that the sender's objective is available in closed form. Our framework
needs none of this: the rule may be uncertain and enters only through its
evaluations, and an adaptive design policy is learned by simulation for any
prior and likelihood that can be sampled.

In clinical trial design, a Bayesian experimenter designing for a prescribed
frequentist test is a long-standing theme. Assurance, or Bayesian predictive
power, chooses a sample size by averaging the power of the test over a
prior~\citep{spiegelhalter1986predictive, ohagan2005assurance,
desantis2025rethinking}, which is a static instance
of~\eqref{eq:fixed-rule-objective}. \citet{ondra2019optimized} design adaptive
trials for a prescribed test by optimising an interim enrichment decision, and
\citet{rossell2013sequential} optimise when to stop a high-throughput genomics
experiment whose results are reported under a fixed false discovery rate rule.
The latter two are bespoke solutions for one experiment and one rule, whereas
our method applies to all of them unchanged.

\section{NUMERICAL EVALUATION}
\label{sec:experiments}

\subsection{Adaptive Clinical Trials}
\label{sec:clinical-trials}

\looseness=-1
We first consider the two-stage trials of \citet{ondra2019optimized}, for a
targeted therapy that may benefit the whole population $F$ or only a
biomarker-positive subgroup $S$. The design has two parts. Before any data are
observed, the experimenter chooses how many patients to recruit from $S$ and
from its complement $\bar{S}$, each randomised equally to treatment and
control. After this first stage, an interim policy maps the observed test
statistics to an action (stop for futility, continue in $S$ only, or continue in
both subgroups) and, if the trial continues, to the second-stage sample sizes.
A regulator then applies a prespecified test of no effect in $S$ and no average
effect in $F$. Because the test controls the family-wise error rate under any
interim policy~\citep{bauer1994evaluation}, the regulator can fix it in
advance. The experimenter's utility, or negative loss, rewards approval in a population in
proportion to its true benefit, minus trial costs.

\begin{table}[!t]
    \centering
    \caption{Expected utility~(millions of euros; higher is better) of designs
    optimised for one test~(rows) and evaluated under each test~(columns).
    Adaptive entries average \(5\) training seeds. Standard errors of paired
    per-trial differences between adaptive entries and the DP or single-stage
    entries are below \(0.05\). The Bayes-act design, the best single-stage
    design under a Bayes-act regulator, runs no trial.}
    \vspace{1ex}
    \label{tab:clinical-trials-utility}
    \small
    \setlength{\tabcolsep}{4pt}
    \begin{tabular*}{\columnwidth}{@{\extracolsep{\fill}}llrr@{}}
        \toprule
        & & \multicolumn{2}{c}{Test applied} \\
        \cmidrule(l){3-4}
        Design & Designed for & Bonferroni & Simes \\
        \midrule
        Bayes-act design & Bayes act & $0.00$ & $0.00$ \\
        \midrule
        Single stage & Bonferroni & $52.27$ & $53.23$ \\
                     & Simes & $52.22$ & $53.29$ \\
        \midrule
        Adaptive (ours) & Bonferroni & $57.79$ & $57.57$ \\
                       & Simes & $57.59$ & $57.76$ \\
        \midrule               
        Oracle DP & Bonferroni & $57.97$ & $57.63$ \\
        \bottomrule
    \end{tabular*}
    \vspace{-2ex}
\end{table}

\citet{ondra2019optimized} find the optimal designs by dynamic
programming~\citep[DP,][]{bellman1957dynamic}, which needs the test's rejection
probabilities in closed form. For this reason, while they note that more
powerful tests could be used, they restrict the regulator to a conservative
Bonferroni-adjusted test. Our method only needs to simulate the rule, so it is
free of this restriction. To show this, we also apply it to a Simes-based
closed testing procedure~\citep{marcus1976closed, simes1986improved}. For each
test, we train a neural interim policy with REINFORCE~\citep{williams1992reinforce}
and, as with DP, choose among $100$ candidate first stages by simulating the
trial each one leads to~(Appendix~\ref{app:clinical-trials}). Like
\citet{ondra2019optimized}, we also compare with the optimal single-stage
design, which fixes the sample sizes in advance and makes no interim decision.

Under Bonferroni, the learned policy recovers $99.7\%$ of the oracle DP
solution's expected utility and improves on the best single-stage design by
$10\%$~(Table~\ref{tab:clinical-trials-utility}). It is also cheaper to obtain:
training and first-stage selection take under four minutes on a CPU, whereas DP takes 54 minutes
on the same machine. Under Simes, for which no DP solution is available, the policy
trained for that test matches or exceeds every other design in the table. Designing
for the wrong test costs little here, but designing for a Bayes-act regulator is
costly: the best such single-stage design runs no trial, and it is worth nothing
under either test.

\subsection{Persuading a Sceptical Consumer}
\label{sec:destructive-lot-inspection}

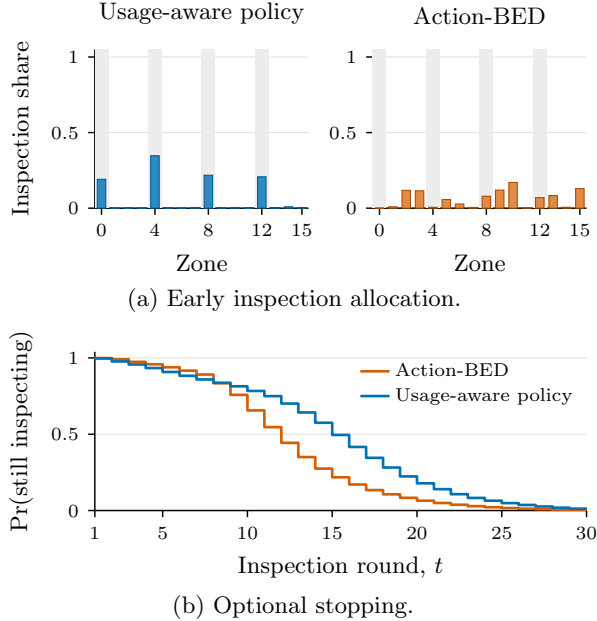
\begin{figure}[t]
  \centering
  \definecolor{ruleawareblue}{RGB}{0,114,178}
\definecolor{actionbedvermillion}{RGB}{213,94,0}
\begin{tikzpicture}
  \begin{groupplot}[
    group style={group size=2 by 2, horizontal sep=0.85cm, vertical sep=1.9cm},
    scale only axis,
    width={0.5*\linewidth-1.3cm}, height=2.1cm,
    axis x line*=bottom,
    axis y line*=left,
    axis line style={draw=black,line width=0.55pt},
    tick align=outside,
    tick style={draw=black,line width=0.55pt},
    major tick length=2.5pt,
    ymajorgrids,
    grid style={draw=black!12,line width=0.35pt},
    tick label style={font=\scriptsize},
    label style={font=\small},
    title style={font=\footnotesize},
    legend style={draw=none, fill=none, font=\scriptsize,
      legend cell align={left}, row sep=-1pt},
    enlargelimits=false,
  ]
    \nextgroupplot[
      title={Usage-aware policy},
      xmin=-0.5, xmax=15.5, xtick={0,4,8,12,15},
      ymin=0, ymax=1.05, ytick={0,0.5,1},
      xlabel={Zone}, ylabel={Inspection share},
      ylabel style={name=ylabtop},
    ]
      \path[fill=black!7] (axis cs:-0.5,0) rectangle (axis cs:0.5,1.05);
      \path[fill=black!7] (axis cs:3.5,0) rectangle (axis cs:4.5,1.05);
      \path[fill=black!7] (axis cs:7.5,0) rectangle (axis cs:8.5,1.05);
      \path[fill=black!7] (axis cs:11.5,0) rectangle (axis cs:12.5,1.05);
      \addplot[ybar, bar width=3.2pt, draw=ruleawareblue!80!black,
        line width=0.4pt, fill=ruleawareblue!82]
        table [x=zone, y=deployment, col sep=comma] {assets/acquisition_map.csv};

    \nextgroupplot[
      title={Action-BED},
      xmin=-0.5, xmax=15.5, xtick={0,4,8,12,15},
      ymin=0, ymax=1.05, ytick={0,0.5,1},
      xlabel={Zone},
    ]
      \path[fill=black!7] (axis cs:-0.5,0) rectangle (axis cs:0.5,1.05);
      \path[fill=black!7] (axis cs:3.5,0) rectangle (axis cs:4.5,1.05);
      \path[fill=black!7] (axis cs:7.5,0) rectangle (axis cs:8.5,1.05);
      \path[fill=black!7] (axis cs:11.5,0) rectangle (axis cs:12.5,1.05);
      \addplot[ybar, bar width=3.2pt, draw=actionbedvermillion!80!black,
        line width=0.4pt, fill=actionbedvermillion!72]
        table [x=zone, y=actionbed, col sep=comma] {assets/acquisition_map.csv};

    \nextgroupplot[
      width={\linewidth-1.75cm},
      at={($(group c1r1.south west)-(0,1.9cm)$)}, anchor=north west,
      xmin=1, xmax=30, ymin=0, ymax=1.04,
      xtick={1,5,10,15,20,25,30},
      ytick={0,0.5,1},
      xlabel={Inspection round, $t$},
      ylabel={Pr(still inspecting)},
      ylabel style={name=ylabbottom},
      legend style={at={(1,1)}, anchor=north east},
      legend image post style={xscale=0.6},
    ]
      \addplot[const plot mark right, mark=none,
        draw=actionbedvermillion, line width=1.0pt,
        line cap=round, line join=round]
        table [x=inspection, y=actionbed, col sep=comma] {assets/stopping_survival.csv};
      \addlegendentry{Action-BED}
      \addplot[const plot mark right, mark=none,
        draw=ruleawareblue, line width=1.0pt,
        line cap=round, line join=round]
        table [x=inspection, y=deployment, col sep=comma] {assets/stopping_survival.csv};
      \addlegendentry{Usage-aware policy}

    \nextgroupplot[group/empty plot, xmin=0, xmax=1, ymin=0, ymax=1]
  \end{groupplot}
  \coordinate (amid) at ($(ylabtop.north)!0.5!(group c2r1.east)$);
  \coordinate (bmid) at ($(ylabbottom.north)!0.5!(group c1r2.east)$);
  \node[anchor=north, font=\small, inner sep=0pt, yshift=-3pt]
    at (amid |- group c1r1.outer south) {(a) Early inspection allocation.};
  \node[anchor=north, font=\small, inner sep=0pt, yshift=-3pt]
    at (bmid |- group c1r2.outer south) {(b) Optional stopping.};
\end{tikzpicture}
  \vspace{-3pt}
  \caption{Accounting for the consumer's rule changes both sampling and
    stopping. (a) Share of the first \(5\) inspections sent to each zone;
    shaded bands mark the zones emphasised by the consumer's prior.
    (b) Fraction of lots still being inspected: the usage-aware policy
    inspects for longer because the consumer requires more evidence. Both
    panels pool \(5\) training seeds with \(10^{4}\) evaluation lots each.}
  \label{fig:lot-policy-behaviour}
\end{figure}

\begin{figure}[!t]
  \centering
  \definecolor{mnistMatched}{HTML}{0072B2}
\definecolor{mnistTransferred}{HTML}{009E73}
\definecolor{mnistStatic}{HTML}{666666}
\definecolor{mnistActionBED}{HTML}{D55E00}
\pgfplotstableread[col sep=comma]{assets/mnist_accuracy.csv}\mnistdata
\begin{tikzpicture}
  \begin{axis}[
    width=\columnwidth,
    height=0.60\columnwidth,
    font=\footnotesize,
    xlabel={Total precision budget \(\Lambda\)},
    ylabel={Test accuracy (\%)},
    xmode=log, log basis x=2,
    xmin=710, xmax=7600, ymin=82, ymax=98,
    xtick={784,1568,3136,6272},
    xticklabels={784,1568,3136,6272},
    ytick={82,86,90,94,98},
    scaled ticks=false,
    axis lines*=left,
    axis line style={line width=0.5pt},
    tick align=outside,
    tick style={line width=0.5pt},
    ymajorgrids=true,
    grid style={gray!25,line width=0.3pt},
    legend style={
      at={(1.0,0.03)}, anchor=south east,
      draw=none, fill=none, font=\footnotesize,
      cells={anchor=west}, inner sep=1pt,
      row sep=1pt,
    },
  ]
    \addplot[name path=matchedupper,draw=none,forget plot]
      table[x=B,y expr=\thisrow{matched}+\thisrow{matchedSD}] {\mnistdata};
    \addplot[name path=matchedlower,draw=none,forget plot]
      table[x=B,y expr=\thisrow{matched}-\thisrow{matchedSD}] {\mnistdata};
    \addplot[mnistMatched,fill opacity=0.13,draw=none,forget plot]
      fill between[of=matchedupper and matchedlower];
    \addplot[name path=transferredupper,draw=none,forget plot]
      table[x=B,y expr=\thisrow{transferred}+\thisrow{transferredSD}] {\mnistdata};
    \addplot[name path=transferredlower,draw=none,forget plot]
      table[x=B,y expr=\thisrow{transferred}-\thisrow{transferredSD}] {\mnistdata};
    \addplot[mnistTransferred,fill opacity=0.13,draw=none,forget plot]
      fill between[of=transferredupper and transferredlower];
    \addplot[name path=staticupper,draw=none,forget plot]
      table[x=B,y expr=\thisrow{static}+\thisrow{staticSD}] {\mnistdata};
    \addplot[name path=staticlower,draw=none,forget plot]
      table[x=B,y expr=\thisrow{static}-\thisrow{staticSD}] {\mnistdata};
    \addplot[mnistStatic,fill opacity=0.13,draw=none,forget plot]
      fill between[of=staticupper and staticlower];
    \addplot[name path=actionbedupper,draw=none,forget plot]
      table[x=B,y expr=\thisrow{actionbed}+\thisrow{actionbedSD}] {\mnistdata};
    \addplot[name path=actionbedlower,draw=none,forget plot]
      table[x=B,y expr=\thisrow{actionbed}-\thisrow{actionbedSD}] {\mnistdata};
    \addplot[mnistActionBED,fill opacity=0.13,draw=none,forget plot]
      fill between[of=actionbedupper and actionbedlower];
    \addplot[mnistMatched,line width=0.7pt,mark=*,mark size=1.6pt]
      table[x=B,y=matched] {\mnistdata};
    \addlegendentry{Usage-aware (ours)}
    \addplot[mnistTransferred,line width=0.7pt,mark=square*,mark size=1.5pt]
      table[x=B,y=transferred] {\mnistdata};
    \addlegendentry{Transferred}
    \addplot[mnistActionBED,line width=0.7pt,mark=diamond*,mark size=2pt]
      table[x=B,y=actionbed] {\mnistdata};
    \addlegendentry{Action-BED}
    \addplot[mnistStatic,line width=0.7pt,mark=triangle*,mark size=1.8pt]
      table[x=B,y=static] {\mnistdata};
    \addlegendentry{Static}
  \end{axis}
\end{tikzpicture}
  \vspace{-6pt}
  \caption{Image acquisition for frozen classifiers. Results average over three
    independently trained classifier pairs, both classifiers within each pair,
    and three policy-training seeds; bands show one standard deviation across
    the three per-pair averages.}
  \label{fig:mnist}
  \vspace{-2ex}
\end{figure}
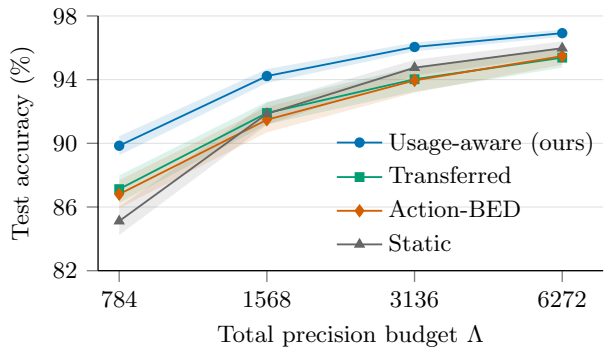

\begin{table}[!t]
    \centering
    \caption{Mean total loss under the sceptical consumer,
    averaged over five training seeds and 10,000 evaluation lots each, with
    the standard deviation across seeds. The loss includes decision error and
    the cost of destroyed units.}
    \vspace{1ex}
    \label{tab:lot-deployment-loss}
    \small
    \begin{tabular}{@{}lr@{}}
        \toprule
        Policy & Loss~($\downarrow$) \\
        \midrule
        Random (full budget) & $0.524 \pm 0.006$ \\
        RL-BOED (full budget) & $0.432 \pm 0.015$ \\
        Action-BED (optional stopping) & $0.648 \pm 0.008$ \\
        Action-BED (forced full budget) & $0.414 \pm 0.022$ \\
        Usage-aware policy~(ours) & $\mathbf{0.265 \pm 0.006}$ \\
        \bottomrule
    \end{tabular}
    \vspace{-2ex}
\end{table}

Our second experiment is a sequential extension of a problem proposed by
\citet{lindley1991evidence}. A manufacturer inspects a lot before a consumer
decides whether to accept it. The lot is divided into $16$ zones, and each
inspection tests a batch of units from one zone, destroying them in the process.
The manufacturer may stop at any point before a budget of $30$ inspections runs
out, and their loss counts wrong decisions as well as destroyed units. The manufacturer and consumer share the same model of how defects arise, but not the same prior or
loss: the consumer believes good lots are rarer, suspects four particular zones,
and accepts only if they are at least $80\%$ sure the lot is good. The manufacturer
must therefore decide where to inspect and when to stop while anticipating how
the consumer's posterior will respond.

The usage-aware policy is a neural network that picks the next zone at
each round. We train it with proximal policy
optimisation~\citep[PPO,][]{schulman2017proximal} directly against the
usage risk, and it stops as soon as the loss of deciding now is no larger
than its critic's estimate of the loss from continuing to inspect. We compare it with
three baselines: Action-BED~\citep{rossa2026actionbed}, which trains a design policy against the
manufacturer's own Bayes risk and stops by the same rule or uses the full budget;
RL-BOED~\citep{blau2022optimizing}, which maximises a lower bound on the EIG;
and random inspection. Full details are in
Appendix~\ref{app:destructive-lot-inspection}.

\looseness=-1
Knowing the consumer's rule changes both where and when the manufacturer
inspects (Figure~\ref{fig:lot-policy-behaviour}). The usage-aware policy
sends $96\%$ of its first five inspections to the four zones the consumer
suspects, where evidence moves the consumer's beliefs the most; Action-BED,
whose own prior favours no zone, sends only $15\%$ there. Action-BED
also stops as soon as its own decision is settled, before the fifteenth
inspection on $73\%$ of the lots. This rarely convinces the consumer, who
accepts fewer than $1\%$ of the lots, so Action-BED does worse than random
inspection. The usage-aware policy instead inspects until the consumer's
decision is settled, and stops that early on only $43\%$ of the lots.
Averaged over five training seeds, its usage risk is $36\%$ below the best
baseline (Action-BED forced to use the full budget) and about half that of
random inspection~(Table~\ref{tab:lot-deployment-loss}).

\subsection{High-Dimensional Image Acquisition}
\label{sec:high-dimens-image}

\begin{figure*}[!t]
    \centering
    \definecolor{componentgray}{gray}{0.28}
\begin{tikzpicture}
    \begin{groupplot}[
        group style={
            group size=3 by 1,
            horizontal sep=1.8cm,
        },
        scale only axis,
        width=4cm,
        height=2.6cm,
        axis x line*=bottom,
        axis y line*=left,
        axis line style={draw=black,line width=0.55pt},
        tick align=outside,
        tick style={draw=black,line width=0.55pt},
        major tick length=2.5pt,
        ymajorgrids,
        grid style={draw=black!12,line width=0.35pt},
        tick label style={font=\small},
        label style={font=\small},
        legend style={
            draw=none,
            fill=none,
            font=\scriptsize,
            cells={anchor=west},
            /tikz/every even column/.append style={column sep=0.8em},
        },
    ]

    \nextgroupplot[
        xlabel={$\kappa$},
        ylabel={Usage risk},
        ylabel style={name=ylab1},
        xmin=-0.1,
        xmax=3.1,
        ymin=0.10,
        ymax=0.64,
        xtick={0,1,1.5,2,2.5,3},
        ytick={0.1,0.2,0.3,0.4,0.5,0.6},
        yticklabels={,0.2,,0.4,,0.6},
        legend style={at={(0.5,1.03)},anchor=south},
        legend columns=3,
    ]
        \addplot[
            black,
            solid,
            line width=1pt,
            mark=triangle*,
            mark size=2.5pt,
            mark options={solid,fill=white,line width=0.7pt},
        ] table[x=kappa,y=kg_star,col sep=comma] {assets/figure3a_risk.csv};
        \addlegendentry{KG}

        \addplot[
            black,
            solid,
            line width=1pt,
            mark=o,
            mark size=2.2pt,
            mark options={solid,fill=white,line width=0.7pt},
        ] table[x=kappa,y=eig,col sep=comma] {assets/figure3a_risk.csv};
        \addlegendentry{EIG}

        \addplot[
            black,
            solid,
            line width=1pt,
            mark=diamond,
            mark size=2.5pt,
            mark options={solid,fill=white,line width=0.7pt},
        ] table[x=kappa,y=random,col sep=comma] {assets/figure3a_risk.csv};
        \addlegendentry{Random}

        \addplot[
            black,
            solid,
            line width=1pt,
            mark=square*,
            mark size=2.1pt,
            mark options={solid,fill=black},
        ] table[x=kappa,y=rule_aware_kg_star,col sep=comma] {assets/figure3a_risk.csv};
        \addlegendentry{\makebox[0pt][l]{Usage-aware (ours)}}

    \nextgroupplot[
        ylabel={Usage risk},
        ylabel style={name=ylab2},
        ymin=0,
        ymax=0.50,
        xmin=-0.55,
        xmax=1.55,
        ytick distance=0.10,
        xtick={0,1},
        xticklabels={KG, Usage-\\aware},
        xticklabel style={align=center,yshift=-1pt},
        yticklabel={\pgfmathprintnumber[fixed,fixed zerofill,precision=2]{\tick}},
        ybar stacked,
        bar width=11mm,
        area legend,
        legend style={at={(0.5,1.03)},anchor=south},
        legend columns=1,
    ]
        \addplot[draw=componentgray,fill=componentgray,line width=0.4pt]
        table[x=index,y=bayes_risk,col sep=comma] {assets/figure3b_decomposition.csv};
        \addlegendentry{Bayes risk $B_{\ell}$}

        \addplot[
            draw=black!65,
            fill=white,
            pattern=north east lines,
            pattern color=black!65,
            line width=0.4pt,
        ]
        table[x=index,y=incompatibility,col sep=comma] {assets/figure3b_decomposition.csv};
        \addlegendentry{Incompatibility $C_{\ell}$}

    \nextgroupplot[
        xmin=-1,
        xmax=17,
        xlabel={Shared candidates},
        xtick={0,4,8,12,16},
        ymin=0,
        ymax=0.1,
        scaled y ticks=false,
        ytick={0,0.02,0.04,0.06,0.08,0.10},
        yticklabels={0,2,4,6,8,10},
        ylabel={Risk reduction ($10^{-2}$)},
        ylabel style={name=ylab3},
        legend style={at={(0.5,1.03)},anchor=south},
        legend columns=2,
    ]
        \addplot[
            black,
            solid,
            line width=1pt,
            mark=*,
            mark size=2.5pt,
            error bars/.cd,
            y dir=both,
            y explicit,
            error bar style={black},
        ] table[x=shared,y=average,y error minus=average_minus,y error plus=average_plus,col sep=comma] {assets/figure3c_two_dm.csv};
        \addlegendentry{Average}

        \addplot[
            black,
            solid,
            line width=1pt,
            mark=square*,
            mark size=2.5pt,
            mark options={solid},
            error bars/.cd,
            y dir=both,
            y explicit,
            error bar style={black,solid},
        ] table[x=shared,y=minimax,y error minus=minimax_minus,y error plus=minimax_plus,col sep=comma] {assets/figure3c_two_dm.csv};
        \addlegendentry{Minimax}

    \end{groupplot}

    \path (current bounding box.south) coordinate (panel captions);
    \foreach \c in {1,2,3} {
        \coordinate (panel mid \c) at ($(ylab\c.north)!0.5!(group c\c r1.east)$);
    }
    \foreach \c/\t in {%
        1/{(a) Usage risk against $\kappa$},
        2/{(b) Risk decomposition at $\kappa=3$},
        3/{(c) Two DMs}%
    } {
        \node[anchor=north, font=\small, inner sep=0pt, yshift=-1ex]
            at (panel mid \c |- panel captions) {\t};
    }
\end{tikzpicture}
    \vspace{-3pt}
    \caption{\looseness=-1 Bayesian ranking and selection. (a) Usage risk
    against the DM's caution $\kappa$. (b) Usage risk split into Bayes risk $B_{\ell}$ and
    incompatibility $C_{\ell}$ at $\kappa=3$. (c) Risk reduction against the
    number of shared candidates for the DMs. \emph{Average}: the average
    policy vs.\ the better specialist on the average
    risk~\eqref{eq:bayes-average-objective}. \emph{Minimax}: the minimax policy
    vs.\ the best randomisation of specialists on the worst-case
    risk~\eqref{eq:minimax-objective}. All panels average $10^{5}$ simulated
    episodes; standard errors are below $0.002$ in (a, b), and $95\%$ paired
    confidence intervals extend at most $0.003$ from the estimates in (c).}
    \label{fig:ranking-and-selection}
    \vspace{-2ex}
\end{figure*}
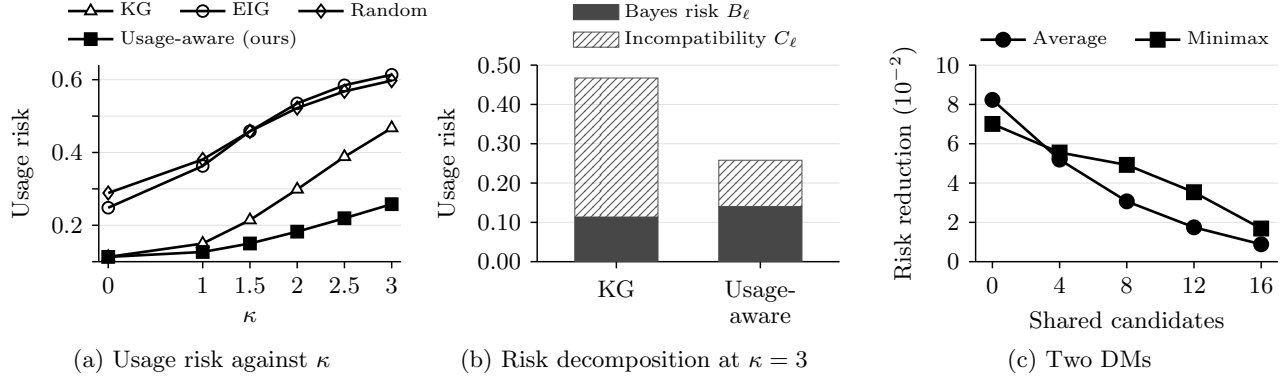

\looseness=-1
We now turn to a learned downstream rule: a classifier
trained in advance and frozen, as in active feature acquisition~\citep{ma2019eddi,
kobayashi2026learning}. The experimenter sees a handwritten MNIST digit only
through a noisy sensor and must decide where to look by splitting a fixed
precision budget $\Lambda$ across the image. The budget is spent in four equal
rounds over $196$ image blocks, so each round's design has $195$ continuous
dimensions. The measurements are passed to a frozen convolutional
classifier~\citep{lecun1989handwritten}, and the downstream loss is its
cross-entropy on the true label. To test how closely the design must be tailored
to the rule, we train two classifiers that differ only in how their training
images were measured: one mostly saw precise measurements of the top of the
digit, the other of the bottom. For each classifier, we train an adaptive
usage-aware policy and compare it with three
baselines~(details in Appendix~\ref{app:high-dimens-image}): the policy trained
for the other classifier (transferred); a static policy trained for the downstream
classifier but with fixed allocations; and Action-BED, which trains the same
adaptive network jointly with its own classifier
via~\eqref{eq:action-bed-formulation}. Since the downstream loss is the log
loss, Action-BED maximises a lower bound on the EIG about the label.

The usage-aware policy is the most accurate at every budget, with the
largest advantage when precision is scarce~(Figure~\ref{fig:mnist}). The
transferred policy and Action-BED perform worse, and once precision is less
scarce, even the static policy overtakes them. As the budget grows, all policies
measure most of the image well and the gaps narrow.

\subsection{Bayesian Ranking and Selection}
\label{sec:ranking-and-selection}

For our final experiment, we perform a controlled study of downstream-rule
mismatch. We use the Bayesian ranking and selection setting of
\citet{frazier2008knowledge}: an experimenter spreads $T=24$ noisy measurements
over $16$ candidates~(e.g., drug compounds) with unknown utilities, after which
a DM deploys one candidate or keeps a default of zero utility. The
experimenter's loss is the utility forgone relative to the best option, so the
Bayes act deploys the candidate with the largest positive posterior mean. Our DM
is more cautious: they consider only candidates whose posterior mean exceeds
zero by more than $\kappa$ posterior standard deviations, so raising $\kappa$
moves the rule continuously away from the Bayes act at $\kappa=0$.

Instead of training a neural policy, we adapt the knowledge
gradient~\citep[KG,][]{frazier2008knowledge, frazier2010paradoxes}, a strong
existing policy for the Bayes act.\footnote{Neural policies trained
on~\eqref{eq:fixed-rule-objective} did not match KG here; see
Appendix~\ref{app:ranking-and-selection}.} KG scores a measurement by the value of
the decision that follows it, so scoring the DM's decision instead makes it
usage-aware. The two policies then differ only in the decision they
anticipate, so any gap between them is due to the rule alone.

The two coincide at $\kappa=0$, but as the DM grows more cautious, the
usage risk of KG rises much faster; EIG maximisation and random sampling do
worse still~(Figure~\ref{fig:ranking-and-selection}a). The usage-aware
policy accepts a somewhat higher Bayes risk to cut the
incompatibility cost by two thirds~(Figure~\ref{fig:ranking-and-selection}b).

Released data are often used by several DMs, and data that serve one may be
useless to another. To study this, we keep the cautious DM above with
$\kappa=3$, and add a second DM who differs in both rule and options. The second
DM instead requires a large estimated utility, deploying their best candidate
with posterior mean above $0.7$. Their action spaces also differ: each DM may
choose from a subset of the $16$ candidates, and we vary how many candidates
they share. A policy tailored to one DM, which we call a specialist, ignores
candidates that only the other can choose. To serve both, we apply KG to a
weighted sum of the values of the two DMs' decisions. Equal weights target the
average objective~\eqref{eq:bayes-average-objective}, and the weight that
minimises the larger of the two DMs' risks targets the minimax
objective~\eqref{eq:minimax-objective}. The average policy beats the better
specialist, and the minimax policy beats the best randomisation between the
specialists, by the widest margins when the DMs share few candidates and each
specialist neglects the other DM's candidates the
most~(Figure~\ref{fig:ranking-and-selection}c). Complete details are in
Appendix~\ref{app:ranking-and-selection}.

\section{DISCUSSION}

We introduced usage-aware experimental design, a framework for choosing
experiments when the data are acted on by external decision makers whose rules
the experimenter cannot change and may not know. These rules may be non-Bayesian
or black-box, several may apply, and the decision
makers' interests may conflict with the experimenter's. We
learn amortised, adaptive design policies by simulation, which needs only
evaluations of the rule and so applies to any rule that can be simulated. The
framework recovers standard BED when the rule is the experimenter's Bayes act,
but strictly generalises it: under some rules, more informative data can hurt,
so no standard BED criterion ranks experiments as the usage risk does. The
value of an experiment thus depends on who acts on its data, and how. We
discuss the limitations of this work and its broader impact in
Appendix~\ref{app:broader-impact}.

\bibliographystyle{plainnat}
\bibliography{references}

\clearpage
\appendix
\thispagestyle{empty}

\onecolumn
\aistatstitle{Supplementary Material for ``Designing Experiments for External
Decision Makers''}

\section{Proofs}
\label{app:proofs}

\subsection{Proof of Proposition~\ref{prop:blackwell-safe-rule}}

Throughout, a~(statistical) experiment is a Markov kernel from $\mathcal{Z}$ to
a data space~\citep{blackwell1953equivalent}. A design policy $\pi_{d}$
induces the experiment $p(\dd D_{T} \mid z; \pi_{d})$, and the usage risk, the Bayes risk, and the Blackwell order depend on the policy only
through this kernel. The proposition is stated over all experiments, and we
keep the notation $\pi_{d}^{j}$ and $D_{T}^{j}$ for a generic experiment and
its data. For any experiment and any prior, the posterior
$b_{T}^{j} \coloneq p_{z}(\cdot \mid D_{T}^{j})$ is a random element of
$\mathcal{P}_{z}$, and by~\eqref{eq:fixed-rule-risk}
and~\eqref{eq:rule-induced-posterior-risk},
\begin{equation}
    R_{\ell}(\pi_{d}^{j}, \pi_{a}) = \mathbb{E}\bigl[r_{\pi_{a}}(b_{T}^{j})\bigr].
    \label{eq:app-risk-as-expected-posterior-risk}
\end{equation}

\paragraph{Sufficiency.}
Suppose that $r_{\pi_{a}}$ is concave and that
$\pi_{d}^{1} \succeq_{\mathrm{B}} \pi_{d}^{0}$ with garbling $G$. Couple
the two experiments by drawing $z$ from the prior, then
$D_{T}^{1} \sim p(\cdot \mid z; \pi_{d}^{1})$, and finally
$D_{T}^{0} \sim G(\cdot \mid D_{T}^{1})$. Since $G$ does not depend on $z$,
the variables $z \to D_{T}^{1} \to D_{T}^{0}$ form a Markov chain, and by
Definition~\ref{def:blackwell} the joint law of $(z, D_{T}^{0})$ under this
coupling is the one induced by $\pi_{d}^{0}$. For each $z \in \mathcal{Z}$, the
tower property and the Markov property give
\begin{equation}
    b_{T}^{0}(z)
    = \mathbb{P}(z \mid D_{T}^{0})
    = \mathbb{E}\bigl[\mathbb{P}(z \mid D_{T}^{1}, D_{T}^{0}) \bigm| D_{T}^{0}\bigr]
    = \mathbb{E}\bigl[\mathbb{P}(z \mid D_{T}^{1}) \bigm| D_{T}^{0}\bigr]
    = \mathbb{E}\bigl[b_{T}^{1}(z) \bigm| D_{T}^{0}\bigr],
\end{equation}
that is, $b_{T}^{0} = \mathbb{E}[b_{T}^{1} \mid D_{T}^{0}]$. Because
$\mathcal{Z}$ is finite, $b_{T}^{1}$ is a bounded random vector in the
simplex $\mathcal{P}_{z}$, on which $r_{\pi_{a}}$ is concave, so the
conditional Jensen inequality gives
$r_{\pi_{a}}(b_{T}^{0}) \geq \mathbb{E}[r_{\pi_{a}}(b_{T}^{1}) \mid D_{T}^{0}]$.
Taking expectations and
using~\eqref{eq:app-risk-as-expected-posterior-risk},
\begin{equation}
    R_{\ell}(\pi_{d}^{0}, \pi_{a})
    = \mathbb{E}\bigl[r_{\pi_{a}}(b_{T}^{0})\bigr]
    \geq \mathbb{E}\bigl[r_{\pi_{a}}(b_{T}^{1})\bigr]
    = R_{\ell}(\pi_{d}^{1}, \pi_{a}).
\end{equation}
This proves sufficiency; note that full support was not used.

\paragraph{Necessity.}
Suppose that $r_{\pi_{a}}$ is not concave. Then there exist beliefs
$b_{1}, b_{2} \in \mathcal{P}_{z}$ and $\lambda \in (0, 1)$ such that,
with $\bar b \coloneq \lambda b_{1} + (1 - \lambda) b_{2}$,
\begin{equation}
    r_{\pi_{a}}(\bar b)
    < \lambda r_{\pi_{a}}(b_{1}) + (1 - \lambda) r_{\pi_{a}}(b_{2}).
    \label{eq:app-nonconcave}
\end{equation}
We exhibit a pair of Blackwell-ordered experiments for which the inequality
of the proposition fails under the prior $p(z)$. We follow the splitting
construction of \citet[Proposition~1]{kamenica2011bayesian}, which shows
that every distribution of posteriors with mean $p$ is induced by some
experiment; we need only a distribution with three atoms and verify it
directly.

The prior need not equal $\bar b$, so we first write it as a mixture that
places mass on $\bar b$. Let $q \coloneq \min_{z} p(z)$ and
\begin{equation}
    b' \coloneq \frac{p - q \bar b}{1 - q},
    \qquad\text{so that}\qquad
    p = q \bar b + (1 - q) b'.
    \label{eq:app-residual-belief}
\end{equation}
Here $q > 0$ because $p$ has full support, and $q < 1$ because
$\mathcal{Z}$ has at least two states, as otherwise $\mathcal{P}_{z}$ is a
single point and $r_{\pi_{a}}$ is trivially concave. Moreover
$b' \in \mathcal{P}_{z}$: its entries sum to one, and they are
non-negative because $p(z) \geq q \geq q \bar b(z)$ for every $z$.

Let $\pi_{d}^{1}$ be the experiment with data space $\{1, 2, 3\}$ and kernel
\begin{equation}
    p(y = 1 \mid z) = \frac{q \lambda b_{1}(z)}{p(z)},
    \qquad
    p(y = 2 \mid z) = \frac{q (1 - \lambda) b_{2}(z)}{p(z)},
    \qquad
    p(y = 3 \mid z) = \frac{(1 - q) b'(z)}{p(z)},
    \label{eq:app-splitting-kernel}
\end{equation}
which is a valid kernel because the numerators are non-negative and sum to
$q \bar b(z) + (1 - q) b'(z) = p(z)$. By Bayes' rule,
\begin{equation}
    p(y = 1)
    = \sum_{z} p(z) \, p(y = 1 \mid z)
    = q \lambda \sum_{z} b_{1}(z)
    = q \lambda,
    \qquad
    p(z \mid y = 1)
    = \frac{p(z) \, p(y = 1 \mid z)}{p(y = 1)}
    = b_{1}(z),
\end{equation}
and likewise $p(y = 2) = q (1 - \lambda)$ with $p(\cdot \mid y = 2) = b_{2}$
and $p(y = 3) = 1 - q$ with $p(\cdot \mid y = 3) = b'$. Let $\pi_{d}^{0}$ be
the experiment that reports only whether $y = 3$, that is, the garbling of
$\pi_{d}^{1}$ that merges outcomes $1$ and $2$. This merge is a Markov
kernel that does not depend on $z$, so
$\pi_{d}^{1} \succeq_{\mathrm{B}} \pi_{d}^{0}$ by
Definition~\ref{def:blackwell}. Under $\pi_{d}^{0}$, the merged outcome has
probability $q \lambda + q (1 - \lambda) = q$ and posterior
\begin{equation}
    p(z \mid y \in \{1, 2\})
    = \frac{p(z) \bigl[p(y = 1 \mid z) + p(y = 2 \mid z)\bigr]}{q}
    = \lambda b_{1}(z) + (1 - \lambda) b_{2}(z)
    = \bar b(z),
\end{equation}
while the outcome $y = 3$ has probability $1 - q$ and posterior $b'$ as
before. By~\eqref{eq:app-risk-as-expected-posterior-risk}, the two
usage risks are therefore
\begin{align}
    R_{\ell}(\pi_{d}^{1}, \pi_{a})
    &= q \lambda \, r_{\pi_{a}}(b_{1}) + q (1 - \lambda) \, r_{\pi_{a}}(b_{2})
    + (1 - q) \, r_{\pi_{a}}(b'), \nonumber \\
    R_{\ell}(\pi_{d}^{0}, \pi_{a})
    &= q \, r_{\pi_{a}}(\bar b) + (1 - q) \, r_{\pi_{a}}(b'),
\end{align}
and subtracting, the contribution of $b'$ cancels, so
by~\eqref{eq:app-nonconcave},
\begin{equation}
    R_{\ell}(\pi_{d}^{1}, \pi_{a}) - R_{\ell}(\pi_{d}^{0}, \pi_{a})
    = q \bigl[\lambda r_{\pi_{a}}(b_{1}) + (1 - \lambda) r_{\pi_{a}}(b_{2})
    - r_{\pi_{a}}(\bar b)\bigr]
    > 0.
\end{equation}
Hence the inequality of the proposition fails under the prior $p(z)$, which
proves necessity. \qed

Full support is used only for necessity. Without it, the proposition holds
with $\mathcal{P}_{z}$ replaced by the simplex over the support of the
prior, which the posterior never leaves: sufficiency applies Jensen's
inequality on that simplex, and necessity runs the construction above with
$\mathcal{Z}$ replaced by the support of the prior.

\subsection{Proof of Corollary~\ref{cor:epu-representation}}

Take the experiments $\pi_{d}^{0}$ and $\pi_{d}^{1}$
constructed in the second half of the proof of
Proposition~\ref{prop:blackwell-safe-rule}, so that
$\pi_{d}^{1} \succeq_{\mathrm{B}} \pi_{d}^{0}$ and
$R_{\ell}(\pi_{d}^{1}, \pi_{a}) > R_{\ell}(\pi_{d}^{0}, \pi_{a})$. Now fix an
action space $\mathcal{A}'$ and a loss
$\ell' \colon \mathcal{A}' \times \mathcal{Z} \to (-\infty, +\infty]$; the
value $+\infty$ is allowed so that the log score is covered, and every
expectation below is then well defined in $(-\infty, +\infty]$. For every
$a \in \mathcal{A}'$,
\begin{equation}
    \mathbb{E}_{\bar b(z)}[\ell'(a, z)]
    = \lambda \mathbb{E}_{b_{1}(z)}[\ell'(a, z)]
    + (1 - \lambda) \mathbb{E}_{b_{2}(z)}[\ell'(a, z)]
    \geq \lambda \mathbb{H}_{\ell'}[b_{1}]
    + (1 - \lambda) \mathbb{H}_{\ell'}[b_{2}],
\end{equation}
and taking the infimum over $a \in \mathcal{A}'$ gives
$\mathbb{H}_{\ell'}[\bar b] \geq \lambda \mathbb{H}_{\ell'}[b_{1}]
+ (1 - \lambda) \mathbb{H}_{\ell'}[b_{2}]$. Since the posterior under $\pi_{d}^{0}$ is
$\bar b$ with probability $q$ and $b'$ with probability $1 - q$, and the
posterior under $\pi_{d}^{1}$ is $b_{1}$, $b_{2}$ and $b'$ with probabilities
$q \lambda$, $q (1 - \lambda)$ and $1 - q$,
\begin{equation}
    B_{\ell'}(\pi_{d}^{1})
    = q \lambda \mathbb{H}_{\ell'}[b_{1}]
    + q (1 - \lambda) \mathbb{H}_{\ell'}[b_{2}]
    + (1 - q) \mathbb{H}_{\ell'}[b']
    \leq q \mathbb{H}_{\ell'}[\bar b]
    + (1 - q) \mathbb{H}_{\ell'}[b']
    = B_{\ell'}(\pi_{d}^{0}).
\end{equation}
Thus the usage risk under $\ell$ strictly prefers $\pi_{d}^{0}$, whereas
the Bayes risk under every loss $\ell'$ weakly prefers $\pi_{d}^{1}$, so no
Bayes risk induces the same ordering as the usage risk on any class of
experiments containing both.
\qed

\section{Experimental Details}
\label{app:experiments}

This appendix gives the details of the experiments in
Section~\ref{sec:experiments}. All experiments are implemented in
JAX~\citep{jax2018github}, and all random seeds, together with any implementation details not stated
here, can be found in the code at
\url{https://codeberg.org/Sahel13/usage-aware-experimental-design}. In every
experiment, the methods are compared on the same held-out test data, which are
used neither for training nor for any selection, such as of checkpoints or
weights, so all comparisons are paired.

\subsection{Adaptive Clinical Trials}
\label{app:clinical-trials}

Confirmatory clinical trials are a natural setting in which the experimenter
does not make the final decision. A sponsor designs the trial, but whether a
therapy is approved, and for whom, is decided by a regulator who applies a
hypothesis test fixed before the trial begins. The test is chosen to control
the family-wise error rate, not to maximise the sponsor's utility, so it need
not resemble the Bayes act under the sponsor's beliefs. We use the two-stage
adaptive trials of \citet{ondra2019optimized}, who optimise two-stage designs
for a targeted therapy that may benefit the whole population or only a
biomarker-positive subgroup, under a prespecified test. They solve the design
problem exactly by dynamic programming, which gives us an optimal adaptive
design against which to measure our method. The DP, however, needs the test's
rejection probabilities in closed form, so they restrict the regulator to a
conservative Bonferroni test. Since our method only simulates the test, we can
also design for a Simes-based test, for which no exact solution is available. We use their societal utility, weak biomarker prior, subgroup
prevalence $\lambda=1/2$ and first cost scenario, in which the biomarker is
costly. Below we describe the model, the designs, the regulator's tests and
the utility, then the learned design, the reference designs and the
evaluation. All sample sizes are per treatment arm.

Patient outcomes are normal with known variance $\sigma^{2}=1$, and the state
is the pair of treatment effects $z=(\delta_{S},\delta_{\bar{S}})$ in the
biomarker-positive subgroup $S$ and its complement $\bar{S}$, with average
effect $\delta_{F}=\lambda\delta_{S}+(1-\lambda)\delta_{\bar{S}}$ in the full
population $F$. The prior puts mass $0.2$, $0.2$, $0.3$ and $0.3$ on
$(0,0)$, $(0.3,0)$, $(0.3,0.15)$ and $(0.3,0.3)$, so the therapy is
ineffective, benefits only $S$, or benefits both subgroups with probabilities
$0.2$, $0.2$ and $0.6$. Stage $k\in\{1,2\}$ recruits $n_{i}^{(k)}$ patients
from subgroup $i\in\{S,\bar{S}\}$, with $n^{(k)}=n_{S}^{(k)}+n_{\bar{S}}^{(k)}$
and trial prevalence $\psi^{(k)}=n_{S}^{(k)}/n^{(k)}$, and patients are
randomised equally to treatment and control. The stage-wise statistic
$Z_{i}^{(k)}$ is the difference in mean outcome between treatment and control
among the stage-$k$ patients of subgroup $i$, divided by its standard error
$\sqrt{2\sigma^{2}/n_{i}^{(k)}}$. Conditional on $z$, the data from earlier
stages, and the stage-$k$ sample sizes, the statistics for the recruited
subgroups are independent with
\begin{equation}
    Z_{i}^{(k)}\sim\mathcal{N}\bigl(\delta_{i}\sqrt{n_{i}^{(k)}/2},\,1\bigr),
    \qquad i\in\{S,\bar{S}\},\quad n_{i}^{(k)}>0,
\end{equation}
and, when both subgroups are recruited, the full-population statistic reweights them by the population
prevalence~\citep[Eq.~2]{ondra2019optimized},
\begin{equation}
    Z_{F}^{(k)}=\kappa^{(k)}\biggl(\frac{\lambda Z_{S}^{(k)}}{\sqrt{\psi^{(k)}}}
    +\frac{(1-\lambda)Z_{\bar{S}}^{(k)}}{\sqrt{1-\psi^{(k)}}}\biggr),
\end{equation}
where $\kappa^{(k)}=\{\lambda^{2}/\psi^{(k)}+(1-\lambda)^{2}/(1-\psi^{(k)})\}^{-1/2}$,
so that $Z_{F}^{(k)}$ has unit variance and zero mean exactly when
$\delta_{F}=0$.

Before any data are observed, the first stage recruits from both subgroups,
with $n_{S}^{(1)}$ and $n_{\bar{S}}^{(1)}$ each taken from the grid
$\{25\times1.3^{j}\colon j=0,\ldots,9\}$, which runs from $25$ to $265.1$.
After the first stage, an interim policy maps the observed statistics to one
of three actions: stop for futility, with $n_{S}^{(2)}=n_{\bar{S}}^{(2)}=0$;
continue in $S$ only, with $n_{S}^{(2)}\in[25,500]$ and $n_{\bar{S}}^{(2)}=0$;
or continue in both subgroups, with
$n_{S}^{(2)},n_{\bar{S}}^{(2)}\in[25,500]$. Following
\citet{ondra2019optimized}, sample sizes are treated as continuous. We also
consider single-stage designs, which fix their sample sizes before any data
are observed, recruit them in one stage with no interim decision, and are
tested once; they recruit $n_{S}\in[50,765]$ and either
$n_{\bar{S}}\in[50,765]$ or $n_{\bar{S}}=0$. Running no trial is always
possible.

The regulator tests $H_{S}\colon\delta_{S}\leq0$ and
$H_{F}\colon\delta_{F}\leq0$ with one of two prespecified tests, both of which
control the family-wise error rate at one-sided level $\alpha=0.025$ under any
interim policy~\citep{bauer1994evaluation}. Write $b_{q}=\Phi^{-1}(1-q)$ and
$\gamma=0.3$. Both tests combine the stages with equal weights,
$Z_{i}^{(c)}=(Z_{i}^{(1)}+Z_{i}^{(2)})/\sqrt{2}$ for $i\in\{S,\bar{S},F\}$, and
reject $H_{F}$ only if the consistency condition $Z_{S}^{(c)}\geq b_{\gamma}$
and $Z_{\bar{S}}^{(c)}\geq b_{\gamma}$ holds. After a futility stop nothing is
rejected, and after continuing in $S$ only $H_{F}$ is not rejected. The
Bonferroni test of \citet{ondra2019optimized} rejects $H_{S}$ if
$Z_{S}^{(c)}\geq b_{\alpha/2}$, and $H_{F}$ if $Z_{F}^{(c)}\geq b_{\alpha/2}$.
The second test is a closed testing procedure~\citep{marcus1976closed} that
tests the intersection $H_{S}\cap H_{F}$ with Simes'
test~\citep{simes1986improved}. With stage-wise $p$-values
$p_{i}^{(k)}=1-\Phi(Z_{i}^{(k)})$ for $i\in\{S,F\}$, the stage-wise Simes
$p$-value of this intersection is
\begin{equation}
    p_{SF}^{(k)}=\min\bigl\{2\min(p_{S}^{(k)},p_{F}^{(k)}),\,
    \max(p_{S}^{(k)},p_{F}^{(k)})\bigr\},
\end{equation}
except that $p_{SF}^{(2)}=p_{S}^{(2)}$ after continuing in $S$ only. With
$C_{SF}=\{\Phi^{-1}(1-p_{SF}^{(1)})+\Phi^{-1}(1-p_{SF}^{(2)})\}/\sqrt{2}$, the
test rejects $H_{S}$ if $C_{SF}\geq b_{\alpha}$ and
$Z_{S}^{(c)}\geq b_{\alpha}$, and $H_{F}$ if $C_{SF}\geq b_{\alpha}$ and
$Z_{F}^{(c)}\geq b_{\alpha}$. Single-stage designs use the same tests with the
single-stage statistics in place of $Z^{(c)}$ and the single-stage Simes
$p$-value compared directly with $\alpha$, and a single-stage design with
$n_{\bar{S}}=0$ tests $H_{S}$ alone at level $\alpha$.

The experimenter's utility is that of \citet{ondra2019optimized}, which
rewards approval in a population in proportion to its benefit to future
patients. With rejection indicators $\phi_{S},\phi_{F}\in\{0,1\}$, approval
yields
\begin{equation}
    G=\begin{cases}
        rN(\delta_{F}-0.1), & \phi_{F}=1,\\
        rN\lambda(\delta_{S}-0.1), & \phi_{S}=1,\ \phi_{F}=0,\\
        0, & \text{otherwise},
    \end{cases}
\end{equation}
where $rN=10^{9}$ euros~\citep[Eq.~5]{ondra2019optimized}, so approving a
population whose effect is below $0.1$ has negative value. The trial costs
\begin{equation}
  \begin{aligned}
    C &= c_{\mathrm{set}}+c_{\mathrm{bio}}\\
      &\quad+2\sum_{k:\,n^{(k)}>0}n^{(k)}\Bigl(c_{\mathrm{pat}}
      +c_{\mathrm{scr}}\max\Bigl\{\frac{\psi^{(k)}}{\lambda},
      \frac{1-\psi^{(k)}}{1-\lambda}\Bigr\}\Bigr),
  \end{aligned}
\end{equation}
with $c_{\mathrm{set}}=10^{6}$, $c_{\mathrm{bio}}=10^{7}$,
$c_{\mathrm{pat}}=5\times10^{4}$ and $c_{\mathrm{scr}}=5\times10^{3}$
euros~\citep[Eqs.~7--8]{ondra2019optimized}. The factor $2$ counts both arms,
the maximum counts the patients who must be screened to enrol a subgroup in a
proportion other than its prevalence, and a stage with $n^{(k)}=0$ costs
nothing. The utility is $U=G-C$. If no trial is run, $C=0$ and neither test
rejects, so $U=0$. In the notation
of Section~\ref{sec:methodology}, the loss is $\ell(a,z)=-G$ and the data cost
is $c(D_{T})=C$.

We now turn to our design. In the notation of Section~\ref{sec:methodology},
the design policy has two parts: the first-stage sizes
$n^{(1)}=(n_{S}^{(1)},n_{\bar{S}}^{(1)})$, and an interim policy $\pi_{\phi}$
that maps $n^{(1)}$ and the first-stage statistics
$Z^{(1)}=(Z_{S}^{(1)},Z_{\bar{S}}^{(1)})$ to the interim action and the
second-stage sizes. Let $V(n^{(1)},\pi_{\phi})$ be the expected utility of
running the first stage $n^{(1)}$ and then following $\pi_{\phi}$. We choose
the two parts in turn. First, we train a single network for all first stages:
each training trial draws $n^{(1)}$ uniformly from the grid, so training
maximises the average of $V(n^{(1)},\pi_{\phi})$ over the $100$ first stages.
Second, with $\pi_{\phi}$ fixed and used deterministically (most probable
action, sizes at the Gaussian means), we estimate $V(n^{(1)},\pi_{\phi})$ for every first stage
on $4\times10^{5}$ simulated trials and select the best, or no trial if none
has positive value. Each first stage is thus judged by the trial it leads to,
as in the backward induction of the DP, and both methods choose from the same
$100$ first stages. Every trained policy selects the DP's first stage,
$(92.8,54.9)$.

The network is a multilayer perceptron with two hidden layers of $64$ SiLU
units, whose inputs are $Z^{(1)}$, the standardised logarithms of $n^{(1)}$,
and the posterior probabilities of the four prior atoms. Its outputs are
logits for the three interim actions and the means of Gaussian distributions
over unconstrained second-stage sizes $u$, one for continuing in $S$ only and
two for continuing in both subgroups, which are mapped to sample sizes by
$n=25+475\,\sigma(u)$ with $\sigma$ the logistic function. The output layer is
initialised at zero, so training starts from a uniform choice of action. The
log standard deviations of the Gaussians are learned parameters, independent
of the input, initialised at $\log0.5$ and clipped to $[-4,0.7]$. In each
training trial, $z$ is drawn from the prior and $Z^{(1)}$ is simulated under
it; to reduce variance, the second stage is then simulated under every prior
atom with shared noise, and its utility is averaged under the posterior given
$Z^{(1)}$. This uses only the experimenter's model, and the regulator's test
is called as a black box. Since the interim action is discrete and the test's
decision is a step function of the statistics, we train with score-function
gradients~\citep{williams1992reinforce}, using the mean reward of
the other trials in the batch that share the same first stage as a baseline,
standardising the advantages, and adding an entropy bonus of $0.01$ on the
interim action. We use
AdamW with weight decay $10^{-5}$, batches of $8192$ trials and gradient
clipping at norm $1$, for up to $16{,}000$ updates. The learning rate rises
linearly to $10^{-3}$ over $400$ updates and then follows a cosine decay to
$10^{-5}$. Every $200$ updates, the deterministic policy is scored on $50{,}000$
fixed validation trials; we keep the best checkpoint and stop once $4{,}000$
updates pass without an improvement of $0.1$ million euros. We set these
hyperparameters by light manual tuning. We train five policies per test, with
different seeds.

We compare with three reference designs. The single-stage designs maximise the
exact expected utility, which under Bonferroni is computed by one-dimensional
quadrature; under Simes, a two-dimensional integrator calls the test as a black
box and locates its rejection boundary by bisection. The search uses a grid of
step $1$ (Bonferroni) or $5$ (Simes), refined locally to step $0.05$, and takes
the best of recruiting from both subgroups, recruiting from $S$ only, and
running no trial. This gives
$(n_{S},n_{\bar{S}})=(203.8,113.0)$ under Bonferroni and $(203.8,104.6)$ under
Simes. The oracle DP follows \citet{ondra2019optimized}, for the Bonferroni
test. For each of the $100$ first stages and each cell of a grid of step $0.1$
on $(Z_{S}^{(1)},Z_{\bar{S}}^{(1)})\in[-6,9.5]^{2}$, it maximises the expected
utility of the remaining trial over the interim action and the second-stage
sizes, using the posterior of $z$ given that the first-stage statistics fall
in that cell and evaluating the test at the cell centre. The interim action and second-stage sizes are searched on a grid of step $12.5$, refined by
golden-section search after continuing in $S$ only and by local grids of steps
$3$ and $0.6$ after continuing in both subgroups. Rejection probabilities are
one-dimensional integrals of closed-form conditional probabilities, computed by
Gauss--Legendre quadrature with $48$ nodes on each side of the integrand's
kink. The DP then selects the first stage with the highest expected utility,
$(n_{S}^{(1)},n_{\bar{S}}^{(1)})=(92.8,54.9)$. Finally, the Bayes-act design is the
single-stage design, or no trial, that would be optimal if the regulator took
the Bayes act under the experimenter's posterior, that is, the one maximising
$\mathbb{E}[\max_{a}\mathbb{E}(G\mid\text{data})]-C$. Under that assumption no
trial is optimal: the prior already favours approval in $F$, worth $87.5$
million euros, against $75.5$ million for the best trial. Under either actual
test, however, no trial means no approval. We did not search two-stage designs
under this assumption, so this comparison is relative to single-stage designs.

Every design in Table~\ref{tab:clinical-trials-utility} is evaluated under both
tests on the same $10^{6}$ test trials, which share the effect pair and the
stage-wise noise across designs and tests; in particular, the DP design,
solved for Bonferroni, is also evaluated under Simes. Learned entries average
the five policies' utilities within each trial. Standard errors of the
paired per-trial differences between our design for each test and the DP and
single-stage designs are below $0.05$ million euros, and the standard
deviation across seeds is at most $0.12$ million euros.
All computations
run on a single Apple M5 Pro CPU. The DP takes $54$ minutes in double
precision, each policy trains in $16$ to $26$ seconds in single precision, and
first-stage selection and evaluation of all designs take under three minutes.

\subsection{Persuading a Sceptical Consumer}
\label{app:destructive-lot-inspection}

In this experiment the DM is Bayesian, but does not share the experimenter's
beliefs or loss. The setting comes from \citet{lindley1991evidence}, who study
acceptance sampling between two adversaries: a manufacturer decides how many
units of a lot to test, after which a consumer, with their own prior and
utility, decides whether to accept the lot. The manufacturer must therefore
collect enough evidence to persuade the consumer, not themselves. We turn their
single sample-size decision into a sequential inspection problem with optional
stopping: the lot is divided into zones, defects cluster around an unknown weak
region, each inspection destroys the units it tests, and the manufacturer
chooses both where to inspect next and when to stop. This lets us test whether
accounting for the DM's rule changes not only which data are collected but also
how much. Below we describe the lot model, the two parties' priors and losses,
the learned policies and baselines, and the evaluation.

A lot consists of $16$ zones arranged on a ring, and its latent state is
$z=(G,J,\rho)$, where $G\in\{0,1\}$ indicates whether the lot is good,
$J\in\{0,\ldots,15\}$ is the centre of the weak region, and $\rho$ is the peak
defect probability there. Inspecting zone $x_{t}$ destroys a batch of $b=8$
units, of which
\begin{equation}
  Y_{t}\mid z,x_{t}\sim\operatorname{Binomial}\bigl(b,p_{x_{t}}(z)\bigr),
  \qquad
  p_{x}(z)=p_{0}+(\rho-p_{0})
  \exp\biggl\{-\frac{d(x,J)^{2}}{2\lambda^{2}}\biggr\},
\end{equation}
are defective, where $d(x,J)=\min\{|x-J|,\,16-|x-J|\}$ is the distance around
the ring, $p_{0}=0.02$ is the defect rate far from the weak region, and
$\lambda=1.5$. A lot is good when $\rho\leq\tau=0.12$: given $G=1$, $\rho$
follows $\mathcal{N}(0.08,0.02^{2})$ truncated to $[p_{0},\tau]$, and given
$G=0$, $\mathcal{N}(0.16,0.025^{2})$ truncated to $(\tau,1]$. The
manufacturer's prior sets $\mathbb{P}(G=1)=0.6$ and makes $J$ uniform. The
manufacturer may inspect at most $T=30$ times, and if they stop after $N\leq T$
inspections and the consumer takes action $a\in\{0,1\}$, with $a=1$ for
acceptance, their loss is the decision error plus the cost of the destroyed
units,
\begin{equation}
  \ell(a,z)+c(D_{N})=\mathds{1}\{a\neq G\}+cbN,
  \qquad c=5\times10^{-4},
\end{equation}
so each inspection costs $cb=0.004$.

The consumer shares the likelihood and the distributions of $\rho$ given $G$,
but is more sceptical. Their prior sets $\mathbb{P}_{\mathrm{DM}}(G=1)=0.4$
and puts $90\%$ of the mass on $J$ equally across the zones $\{0,4,8,12\}$,
spreading the remaining $10\%$ uniformly over all $16$ zones. A false
acceptance costs them four times as much as a false rejection, so their Bayes
act, and hence the rule $\pi_{a}$, accepts exactly when
\begin{equation}
  \mathbb{P}_{\mathrm{DM}}(G=1\mid D_{N})\geq\frac{4}{4+1}=0.8.
\end{equation}
The per-zone inspection and defect counts are sufficient statistics of the
binomial likelihood, so both parties' posteriors are functions of these two
$16$-vectors. We evaluate them on a common grid of $2{,}048$ particles, one for
each quality class, location, and $64$ midpoint quantiles of the
class-conditional distribution of $\rho$, weighted by the respective prior.

The usage-aware policy chooses the next zone with an actor network and
decides when to stop with a critic. Both are fully connected networks with two
hidden ReLU layers of width $128$, and both take as input, for each zone, the
number of inspections divided by $T$, the empirical defect rate and the number
of defects divided by $bT$, together with the fraction of the budget used. The
actor outputs logits over the $16$ zones, with its output layer initialised at
zero so that the initial policy is uniform, and the critic outputs a scalar
estimate of the return from continuing to inspect. We train both with
PPO~\citep{schulman2017proximal} against the usage risk~(Appendix~\ref{app:policy-learning}). Rewards are terminal and
undiscounted, so the advantage of a zone is the episode's reward minus the
critic's value of the state before it. Each of $2{,}000$ updates simulates one
episode for each of $256$ lots drawn from the manufacturer's prior and then
takes four epochs over four minibatches of $64$ episodes, using Adam with
learning rate $2.5\times10^{-4}$, gradients clipped to global norm $0.5$,
clipping parameter $0.2$ for both the policy ratio and the value update, value
coefficient $0.5$, advantages standardised within each minibatch, and no
entropy bonus. We keep the final parameters. Some of these
hyperparameters were set by light manual tuning, without a systematic search,
and the rest are common PPO defaults.

Stopping is not a learned action. The reward of stopping after $t$ inspections
is known in closed form: it is minus the manufacturer's posterior expected loss
when the consumer decides on the current data,
$-\mathbb{P}(G\neq\pi_{a}(D_{t})\mid D_{t})-cbt$. The policy stops as soon as
this reward is at least the critic's value. During training it continues to inspect
anyway with probability $0.05$, so that the critic keeps learning the value of
continuing in states where the rule stops. A fresh critic predicts zero, which
exceeds every stop reward, so early training uses the full budget. We also
tried learning stopping as a seventeenth action with PPO, but because stopping
is wrong in almost every state early in training, its probability collapsed
within a few hundred updates and never recovered.

Action-BED~\citep{rossa2026actionbed} is trained identically but against the
manufacturer's own Bayes risk. Its action is the manufacturer's Bayes act,
which accepts exactly when $\mathbb{P}(G=1\mid D_{t})>1/2$ and is computed on
the grid, so no action network is needed, and its stop reward is
$-\min\{\mathbb{P}(G=1\mid D_{t}),\mathbb{P}(G=0\mid D_{t})\}-cbt$. We also
report an ablation that forces Action-BED to use the full budget.
RL-BOED~\citep{blau2022optimizing} is trained identically with terminal reward
given by the sPCE lower bound on the information gain about $z$, with $1{,}023$
contrastive draws from the manufacturer's prior, minus the cost of the
destroyed units. RL-BOED and random inspection always use all $30$
inspections.

Each learned policy is trained with five seeds, and each seed is evaluated on
$10{,}000$ new lots drawn from the manufacturer's prior. Zones are sampled from the policy, the stopping rule is applied
without exploration, and the consumer then decides by their rule.
Table~\ref{tab:lot-deployment-loss} reports the mean and standard deviation
across seeds of each seed's mean realised loss, and the zone frequencies and
stopping curves in Figure~\ref{fig:lot-policy-behaviour} pool the evaluation
episodes of all five seeds.

Each training run
takes about three minutes on a single NVIDIA H100 GPU, so the $20$ runs take
about an hour, and evaluating all policies takes under a minute.

\subsection{High-Dimensional Image Acquisition}
\label{app:high-dimens-image}

In this experiment the downstream decision maker is not a person or a
hand-specified rule but a learned model: a classifier trained in advance on
historical data and frozen before the experiment. The experimenter cannot change the
classifier and can only decide which data to collect for it. Concretely, the
experimenter observes a handwritten digit from MNIST~\citep{lecun1998gradient}
through a noisy sensor and must decide how to distribute a fixed precision
budget across the image, after which the measurements are passed to a frozen
convolutional neural network~\citep{lecun1989handwritten} that predicts the
digit's label. To test how closely the experiment must be tailored to the
downstream rule, we train two versions of this classifier that are identical
except for how their training images were measured: one mostly saw precise
measurements of the top of the digit, the other of the bottom. Below we
describe, in turn, the sensing environment, the two classifiers, the
acquisition policies and baselines, and the evaluation protocol.

Each episode starts from a digit label $c\in\{0,\ldots,9\}$ and its image,
which we randomly reposition so that the digit does not always occupy the same
pixels. The canvas is shrunk by a scale drawn uniformly from $[0.75,0.9]$ and
shifted, with bilinear interpolation, by a uniformly drawn translation that
keeps the shrunken canvas inside the frame with a one-pixel margin, so the digit
is never clipped. Writing $\theta\in[0,1]^{28\times28}$ for the repositioned
image, the latent state is $z=(\theta,c)$. The placement is independent of the
label, and neither the image, its label nor the placement is an input to the
acquisition policy; labels are used only as training targets and for
evaluation.

The experimenter measures this image with a total precision budget
$\Lambda\in\{784,1568,3136,6272\}$, spent in $T=4$ rounds of
$\Lambda_{t}=\Lambda/4$ each. Since the round budgets are fixed, the policy
decides only where to spend each round's budget, not how much to spend or when
to stop. We partition the $P=784$ pixels into $K=196$ non-overlapping
$2\times2$ blocks $\{\mathcal{P}_{j}\}_{j=1}^{K}$, and the design at round $t$
is an allocation of the round's precision over these blocks,
\begin{equation}
    x_{t}\in\mathcal{X}_{t}(\Lambda)
    \coloneq\Bigl\{x\in[0,\infty)^{K}:\textstyle\sum_{j=1}^{K}x_{j}=\Lambda_{t}\Bigr\},
    \label{eq:mnist-design-space}
\end{equation}
where $x_{t,j}$ is the precision assigned to block $j$. Each round therefore
has $K-1=195$ free continuous dimensions. Within a block, the precision is split
equally among the four pixels, $w_{t,p}=x_{t,j}/4$ for $p\in\mathcal{P}_{j}$,
and pixel $p$ is measured as
\begin{equation}
    Y_{t,p}=\sqrt{w_{t,p}}\,\theta_{p}+\sigma\varepsilon_{t,p},
    \label{eq:mnist-sensor}
\end{equation}
with $\sigma=1$ and $\varepsilon_{t,p}$ independent standard Gaussians across
pixels and rounds. For $w_{t,p}>0$, $Y_{t,p}/\sqrt{w_{t,p}}$ is $\theta_{p}$
plus noise of variance $\sigma^{2}/w_{t,p}$, while zero precision supplies no
information, so the budget controls the quality of the measurements rather
than the number of rounds or images. The precision spent on one image totals
$\sum_{t}\sum_{p}w_{t,p}=\Lambda$, so under uniform sensing each pixel receives
a cumulative precision of $\Lambda/P\in\{1,2,4,8\}$.

Rather than working with the raw measurements, both the classifiers and the
acquisition policy summarise the history after round $t$ through the per-pixel
statistics
\begin{equation}
    W_{t,p}=\sum_{s=1}^{t}w_{s,p},\qquad
    S_{t,p}=\sum_{s=1}^{t}\sqrt{w_{s,p}}\,Y_{s,p},\qquad
    R_{t,p}=\frac{S_{t,p}}{W_{t,p}+\sigma^{2}},\qquad
    C_{t,p}=\log\Bigl(1+\frac{W_{t,p}}{\sigma^{2}}\Bigr),
    \label{eq:mnist-representation}
\end{equation}
with $W_{0,p}=S_{0,p}=0$, and take as input the two-channel image
$(R_{t},C_{t})$. Here $R_{t}$ is a stabilised noisy image and $C_{t}$ records
how reliably each pixel has been measured. Both channels are computed from the
observations and the known allocations alone; in particular, $R_{t}$ is not
assumed to be a posterior mean under the MNIST image distribution.

MNIST is released under the Creative Commons Attribution-ShareAlike 3.0
licence. We partition its $60{,}000$ training images once, by a fixed seed, into
$40{,}000$ images for training the classifiers, $10{,}000$ for training the
acquisition policies and $10{,}000$ for validation, and reserve the $10{,}000$
official test images for the results in Figure~\ref{fig:mnist}. Placements and
sensor noise are drawn afresh for every training batch, whereas the validation
and test images use fixed placements and noise, shared by every model.

We now describe the frozen classifiers that act as downstream decision makers.
We train two ten-class CNNs, indexed by
$u\in\{\mathrm{upper},\mathrm{lower}\}$, with the architecture in
Table~\ref{tab:mnist-architectures}. The two classifiers of a pair share their
initialisation, training images, placements, sensor noise and budgets, and
differ only in the historical sensing distribution that produced their training
inputs. In each round of a training history, the precision follows a smooth
Gaussian bump in the digit's canonical coordinate frame (normalised to
$[-1,1]^{2}$ and mapped through the sampled placement), with centre
$(c_{x},c_{y})$ and widths $s_{x}\sim\mathcal{U}(0.4,0.7)$ and
$s_{y}\sim\mathcal{U}(0.25,0.45)$, normalised over the blocks and mixed with a
$15\%$ uniform component. The horizontal centre is always drawn as
$c_{x}\sim\mathcal{U}(-0.5,0.5)$; the two classifiers differ in the vertical
centre. With probability $0.8$, drawn once per image, the history is biased:
the upper classifier draws $c_{y}\sim\mathcal{U}(-0.6,-0.3)$, which places the
bump over the top of the digit, and the lower classifier draws from its
reflection, $c_{y}\sim\mathcal{U}(0.3,0.6)$. Otherwise, both draw
$c_{y}\sim\mathcal{U}(-0.7,0.7)$. The two histories are thus exact vertical
reflections of each other, and labels play no role in them. Because each
training batch contains equally many images at each of the four budgets, a
single classifier serves every budget. Each classifier minimises the terminal
cross-entropy of $(R_{4},C_{4})$ under its history, with the hyperparameters in
Table~\ref{tab:mnist-hyperparameters}; we keep the parameters with the lowest
validation cross-entropy, averaged over the four budgets, after any epoch, and
then freeze the weights for all policy training and evaluation.

In the notation of Section~\ref{sec:methodology}, write
$F_{u}(k\mid R_{4},C_{4})$ for the frozen probability of class $k$ and
$D_{t}=(x_{s},Y_{s})_{s=1}^{t}$ for the history. The downstream action is the
class-probability vector and the loss is the log loss,
\begin{equation}
    \pi_{a}^{u}(D_{4})=\bigl(F_{u}(k\mid R_{4},C_{4})\bigr)_{k=0}^{9},
    \qquad
    \ell(a,z)=-\log a_{c}.
    \label{eq:mnist-downstream-rule}
\end{equation}
These rules are fixed learned classifiers, not Bayes-optimal responses to the
acquired data.

\begin{table}[t]
    \centering
    \caption{Network architectures. Convolutions are $3\times3$ with unit
    stride and same padding, followed by $2\times2$ pooling of stride~$2$.
    Weights use LeCun-normal initialisation and zero biases, except the policy's
    output layer, which is initialised at zero so that every policy starts
    from the uniform allocation.}
    \vspace{1ex}
    \label{tab:mnist-architectures}
    \small
    \begin{tabular}{@{}llll@{}}
        \toprule
        & Classifier $F_{u}$, $q_{\omega}$ & Adaptive policy $f_{\phi}$ & Static policy \\
        \midrule
        Input & $(R_{4},C_{4})$, $2\times28\times28$ & $(R_{t-1},C_{t-1})$, $2\times28\times28$ & none \\
        Conv 1 & $32$ channels, ReLU, max-pool & $24$ channels, SiLU, average-pool & \\
        Conv 2 & $64$ channels, ReLU, max-pool & $32$ channels, SiLU, average-pool & \\
        Context & & $(t/T,\,b_{t}/\Lambda)$ appended to the flattened features & \\
        Dense 1 & $128$ units, ReLU & $128$ units, SiLU & \\
        Output & $10$ class logits & $196$ block logits & $4\times196$ logits $\eta_{t}$ \\
        Parameters & $421{,}930$ & $233{,}772$ & $784$ \\
        \bottomrule
    \end{tabular}
\end{table}

\begin{table}[t]
    \centering
    \caption{Training hyperparameters. All values are shared across budgets,
    classifiers and seeds. They were set to common defaults without tuning,
    except the policy epoch count, which we raised to $120$ after training
    curves of preliminary runs showed that all policies had converged by
    then. The learning rate follows a cosine decay to zero
    over training. After every epoch the model is scored on the validation
    images and the parameters with the lowest validation cross-entropy are
    kept.}
    \vspace{1ex}
    \label{tab:mnist-hyperparameters}
    \small
    \begin{tabular}{@{}llll@{}}
        \toprule
        & Classifiers & Adaptive policies, Action-BED & Static policies \\
        \midrule
        Training images & $40{,}000$ & $10{,}000$ & $10{,}000$ \\
        Optimiser & AdamW & AdamW & AdamW \\
        Peak learning rate & $10^{-3}$ & $10^{-3}$ & $0.03$ \\
        Weight decay & $10^{-4}$ & $10^{-4}$ & $0$ \\
        Batch size & $256$ & $256$ & $256$ \\
        Epochs & $30$ & $120$ & $120$ \\
        Gradient clipping (global norm) & none & $5$ & $5$ \\
        Budgets per model & all four & one & one \\
        \bottomrule
    \end{tabular}
\end{table}

With the downstream rule fixed, we turn to the experimenter's acquisition
policy. Before round $t$, let $b_{t}=\sum_{s=t}^{4}\Lambda_{s}$ be the remaining
budget. A CNN $f_{\phi}$~(Table~\ref{tab:mnist-architectures}) maps the current
representation $(R_{t-1},C_{t-1})$ and the context $(t/T,b_{t}/\Lambda)$ to $K$
logits, and the round's budget is allocated in proportion to their softmax,
\begin{equation}
    h_{t}=f_{\phi}(R_{t-1},C_{t-1},t,b_{t}),\qquad
    x_{t,j}=\Lambda_{t}\frac{\exp(h_{t,j})}{\sum_{k=1}^{K}\exp(h_{t,k})},
    \label{eq:mnist-policy}
\end{equation}
with each block's share floored at $10^{-30}$ so that the gradient of the
$\sqrt{w}$ weighting stays finite. Since the input before the first round is
all zeros, the first allocation is the same for every image. We train a
separate policy for each frozen classifier and total budget by minimising its
usage risk,
\begin{equation}
    \mathcal{L}_{u,\Lambda}(\pi_{d})
    \coloneq\mathbb{E}\bigl[-\log F_{u}(c\mid R_{4}^{\pi_{d}},C_{4}^{\pi_{d}})\bigr],
    \label{eq:mnist-objective}
\end{equation}
over $\phi$, where the expectation is over the image and label, the placement
and the measurement noise. Gradients propagate through all four sensing rounds
and through the frozen classifier's input, so no reinforcement-learning
optimiser is needed; only the acquisition parameters are updated. The
hyperparameters are again listed in Table~\ref{tab:mnist-hyperparameters}.

We compare this usage-aware policy with three baselines, all of which use
the same round budgets, observation model and final classifier input. The
static policy learns one observation-independent logit vector $\eta_{t}$ per
round, $x_{t}=\Lambda_{t}\operatorname{softmax}(\eta_{t})$, by the same
objective, so the observations still enter the terminal classification but
never the allocation. The transferred policy is the adaptive policy trained for
the other classifier, $\widehat{\pi}_{v,\Lambda}^{\mathrm{ad}}$ with $v\neq u$,
evaluated under $F_{u}$; it changes the acquisition without changing the
downstream classifier, and so compares matched and transferred acquisition
directly.

The third baseline, Action-BED~\citep{rossa2026actionbed}, is adaptive and
information-maximising but agnostic to the downstream classifier. It replaces the
frozen classifier in~\eqref{eq:mnist-objective} by a classifier $q_{\omega}$ of
the same architecture and trains both jointly,
\begin{equation}
    \min_{\phi,\omega}\;
    \mathbb{E}\bigl[-\log q_{\omega}(c\mid R_{4}^{\pi_{d}^{\phi}},C_{4}^{\pi_{d}^{\phi}})\bigr],
    \label{eq:mnist-action-bed}
\end{equation}
which is~\eqref{eq:action-bed-formulation} for the log
loss~\eqref{eq:mnist-downstream-rule}. This objective equals the entropy of the
label, approximately $\log10$ for MNIST, minus the Barber--Agakov lower bound
on the mutual information between the label and the
history~\citep{barber2003im,foster2020unified}, so minimising it maximises a
lower bound on the EIG about the label. The downstream classifiers play no part in
training: $q_{\omega}$ is initialised from an additional classifier trained alongside the pair,
trained like $F_{\mathrm{upper}}$ and $F_{\mathrm{lower}}$ but under uniform
sensing, and checkpoints are selected on the validation cross-entropy of
$q_{\omega}$ itself. To make the comparison as close as possible, for a given
seed the Action-BED design network has the same initialisation, training images,
mini-batches, placements, noise, optimiser and number of epochs as the
usage-aware policy, while $q_{\omega}$ has its own AdamW optimiser with the
same settings, so that clipping one network's gradient does not affect the
other's. The two methods therefore differ only in whether the classifier behind
the policy is frozen or learned. We train one Action-BED policy per classifier
pair, budget and seed, and evaluate it under both downstream classifiers. When
scored by its own classifier $q_{\omega}$ instead, Action-BED reaches $92.6\%$,
$96.0\%$, $97.3\%$ and $97.8\%$ accuracy at the four budgets, above every policy
under the downstream classifiers in Figure~\ref{fig:mnist}. Its weaker downstream
accuracy therefore reflects the mismatch with the downstream rule rather than an
uninformative design policy.

Finally, we evaluate every policy by its terminal classification accuracy under
the downstream classifier,
\begin{equation}
    \operatorname{Acc}_{u,\Lambda}(\pi_{d})
    =\mathbb{P}\Bigl(\argmax_{k}F_{u}(k\mid R_{4}^{\pi_{d}},C_{4}^{\pi_{d}})=c\Bigr),
\end{equation}
on the $10{,}000$ test images with three paired placement and noise replicates.
The whole experiment is repeated for three independently trained classifier
pairs, and within each pair we train three policy seeds per method and, where
applicable, target classifier. Accuracy is averaged within each pair over the
two downstream classifiers and the three seeds; Figure~\ref{fig:mnist} reports the
mean of these three per-pair values, and its bands their standard deviation.

Each training
or evaluation run uses a single GPU, an NVIDIA RTX 6000 Ada or A40. Training
the two downstream classifiers and the additional uniform-sensing classifier
takes a few minutes, training the pair's $60$
policies~(two adaptive and two static targets and Action-BED, at four budgets
and three seeds) takes about $40$ minutes on the RTX 6000 Ada, and evaluating
all $180$ policies takes about $4$ minutes, so the whole experiment runs in
roughly two and a half GPU hours.

\subsection{Bayesian Ranking and Selection}
\label{app:ranking-and-selection}

This experiment is a controlled study of how a mismatch between the downstream
rule and the Bayes act changes the experiment one should run, and of designing
a single experiment for several DMs. We use the Bayesian ranking and selection
problem of~\citet{frazier2008knowledge}: an experimenter spreads a fixed number
of noisy measurements over candidates with unknown utilities, such as drug
compounds in a screening campaign, after which a DM deploys one candidate or
keeps a default. Two features make this setting suitable for a controlled
study. First, the posterior is available in closed form, so a single caution
parameter can move the DM's rule continuously away from the Bayes act. Second,
the knowledge gradient~\citep[KG,][]{frazier2008knowledge,frazier2010paradoxes}
is a strong, well-studied policy for the Bayes act in this model, and it only
needs to evaluate the final decision. Valuing the DM's decision in place of the
Bayes act therefore makes it usage-aware, and the two KG policies differ
only in the decision they anticipate, so any gap between them is due to the rule
alone. Below we describe the model, the DMs' rules and the usage risk, the
policies, and their evaluation, first for one DM and then for two.

Candidate $i\in\{1,\ldots,d\}$ has unknown utility
$Z_{i}\overset{\mathrm{iid}}{\sim}\mathcal{N}(\mu,\tau^{2})$, and the default,
indexed $0$, has known utility $Z_{0}=0$. We use $d=16$, $\tau=1$ and
$\mu=\Phi^{-1}(2/16)$, where $\Phi$ is the standard normal distribution
function, so that in expectation two candidates beat the default. In each of
$T=24$ rounds the policy measures one candidate $x_{t}$ and observes
$y_{t}=Z_{x_{t}}+\epsilon_{t}$, with
$\epsilon_{t}\overset{\mathrm{iid}}{\sim}\mathcal{N}(0,\sigma^{2})$ and
$\sigma=0.75$. If candidate $i$ has been measured $n_{t,i}$ times after round
$t$, with observations summing to $q_{t,i}$, its posterior is
$\mathcal{N}(m_{t,i},s_{t,i}^{2})$ with
$s_{t,i}^{2}=(\tau^{-2}+n_{t,i}\sigma^{-2})^{-1}$ and
$m_{t,i}=s_{t,i}^{2}(\tau^{-2}\mu+\sigma^{-2}q_{t,i})$.

The experimenter's loss is the simple regret,
$\ell(a,Z)=\max\{0,Z_{1},\ldots,Z_{d}\}-Z_{a}$, so the Bayes act deploys the
candidate with the largest positive posterior mean. The DMs we consider instead
first screen the candidates, keeping those in
\begin{equation}
    \mathcal{E}_{\kappa,\delta}(D_{t})=\{i:m_{t,i}-\kappa s_{t,i}>\delta\},
    \label{eq:rs-eligible}
\end{equation}
and then deploy the eligible candidate with the largest posterior mean, keeping
the default if none is eligible. In the single-DM study of
Figure~\ref{fig:ranking-and-selection}(a, b), $\delta=0$ and
$\kappa\in\{0,1,1.5,2,2.5,3\}$. Such a DM deploys only candidates with
$\mathbb{P}(Z_{i}>0\mid D_{T})>\Phi(\kappa)$. This eligibility threshold is the
Bayes rule for a binary certification loss that charges $\Phi(\kappa)$ for
certifying a harmful candidate and $1-\Phi(\kappa)$ for rejecting a beneficial
one. At $\kappa=0$, the complete deployment rule is the Bayes act for $\ell$.
Writing $a_{\kappa}(D_{T})$ for the DM's action,
the deployed utility has conditional expectation $m_{T,a_{\kappa}(D_{T})}$, with
$m_{T,0}=0$, so the usage risk is
$R_{\ell}^{\kappa}=\mathbb{E}[\max\{0,Z_{1},\ldots,Z_{d}\}]
-\mathbb{E}[m_{T,a_{\kappa}(D_{T})}]$. The first term, $0.6402$, is the same for
every policy and is computed by quadrature. In the
decomposition~\eqref{eq:deployment-risk-decomposition}, the Bayes risk
$B_{\ell}$ replaces $m_{T,a_{\kappa}(D_{T})}$ by the Bayes value
$\max\{0,\max_{i}m_{T,i}\}$, and the incompatibility is
$C_{\ell}^{\kappa}=R_{\ell}^{\kappa}-B_{\ell}$.

Both KG policies value a measurement by its expected effect on a terminal value
of the posterior. The usage-aware policy uses the deployed utility
$V_{\kappa,\delta}(D_{t})=m_{t,a_{\kappa}(D_{t})}$, and the KG baseline uses the
Bayes value $V_{0,0}(D_{t})=\max\{0,\max_{i}m_{t,i}\}$. Measuring candidate $i$
a further $n$ times leaves the other posteriors unchanged, sets its posterior
standard deviation to $s_{i}'=(s_{t,i}^{-2}+n\sigma^{-2})^{-1/2}$, and moves its
posterior mean to $m_{t,i}+\eta_{i}\xi$, where
$\eta_{i}=(s_{t,i}^{2}-{s_{i}'}^{2})^{1/2}$ and $\xi\sim\mathcal{N}(0,1)$. Let
$b_{i}$ be the largest posterior mean among the other candidates in
$\mathcal{E}_{\kappa,\delta}(D_{t})$, or zero if there is none. Candidate $i$ is
then deployed exactly when its new mean exceeds
$c_{i}=\max\{b_{i},\delta+\kappa s_{i}'\}$, so the expected terminal value is
\begin{equation}
    Q_{i}(n)=b_{i}+(m_{t,i}-b_{i})\Phi(u_{i})+\eta_{i}\varphi(u_{i}),
    \qquad
    u_{i}=\frac{m_{t,i}-c_{i}}{\eta_{i}},
    \label{eq:rs-kg}
\end{equation}
with $\varphi$ the standard normal density. For a cautious DM a single
measurement rarely lifts a candidate over the hurdle $\kappa s_{i}'$, so
one-step scores are nearly flat. We therefore follow the multi-step variant
KG$(*)$ of~\citet{frazier2010paradoxes}, which we call KG in the main text: each
policy measures the candidate with the largest value per measurement,
$\max_{1\leq n\leq T-t}\{(Q_{i}(n)-V_{\kappa,\delta}(D_{t}))/n\}$, and
recomputes the scores after every observation. For $n=1$, $\kappa=0$ and
$\delta=0$ this is the knowledge gradient of~\citet{frazier2008knowledge}, with
the default among the alternatives. We also compare with two rule-agnostic
baselines. For a fixed allocation of measurement counts, the EIG about $Z$ is
$\frac{1}{2}\sum_{i}\log(1+n_{T,i}\tau^{2}/\sigma^{2})$, which does not depend
on the observations and is maximised by spreading the measurements as equally as possible, so
the EIG policy measures the candidates in turn, in a random order. Random
sampling measures a uniformly random candidate in each round.

We also tried learning neural design policies directly on the usage risk~\eqref{eq:fixed-rule-objective}, as in the other experiments, with both
REINFORCE~\citep{williams1992reinforce} and
PPO~\citep{schulman2017proximal}. None of them matched the usage-aware
KG$(*)$ policy, so we use KG$(*)$ instead. This does not weaken the comparison:
KG$(*)$ is a strong policy for this conjugate Gaussian model, and making it
usage-aware requires only evaluations of the DM's rule, so it plays the
same role as a learned policy while holding the design method fixed across
rules.

We run every policy on the same $100{,}000$ simulated episodes. Each episode
fixes the utilities and every potential observation in advance, the $n$th
measurement of a candidate using its $n$th noise draw, so that policies are
compared on common random numbers even when their measurements differ. The
usage-aware policy is run once per $\kappa$, whereas the other policies do
not depend on the rule and are run once and scored under every $\kappa$.
Figure~\ref{fig:ranking-and-selection}(a) reports the episode averages of
$R_{\ell}^{\kappa}$, and Figure~\ref{fig:ranking-and-selection}(b) those of
$B_{\ell}$ and $C_{\ell}^{3}$.

In the second study we keep the cautious DM above with $\kappa=3$ and
$\delta=0$, now called DM1, and add a DM2 with $\kappa=0$ and $\delta=0.7$, which
deploys its best candidate with posterior mean above $0.7$. The two DMs also
differ in their options: each may deploy $(16+k)/2$ of the candidates, $k$ of
which are shared, so their options are disjoint at $k=0$ and identical at
$k=16$, and we vary $k\in\{0,4,8,12,16\}$. For a weight $w\in[0,1]$, the policy
applies KG$(*)$ to the terminal value $wV_{1}+(1-w)V_{2}$, where $V_{j}$ is
DM~$j$'s deployed utility under its own rule and options. By linearity, its
score for $n$ further measurements is the weighted sum of the two DMs' scores,
with each DM's score set to zero for a candidate it may not deploy. Otherwise,
the score is computed as above with $b_{i}$ taken over the other eligible
candidates that DM may deploy. The maximum over $n$ is taken after summing.
The specialists are $w=1$ and $w=0$, and the average
policy, for the objective~\eqref{eq:bayes-average-objective} with equal weight
on the two DMs, is $w=1/2$. The minimax
objective~\eqref{eq:minimax-objective} does not decompose over measurements, so
the minimax policy instead selects its weight by validation: the
$w\in\{0,0.05,\ldots,1\}$ that maximises $\min\{J_{1},J_{2}\}$ on $20{,}000$
validation episodes, where $J_{j}$ is DM~$j$'s expected deployed utility, with
near-ties within $10^{-4}$ resolved towards $w=1/2$. We compare it with the
best randomisation between the specialists, which follows one of them for a
whole episode with the probability that maximises $\min\{J_{1},J_{2}\}$ on the
same episodes. The $100{,}000$ test episodes are drawn as two independent sets of
$50{,}000$. Since
each DM's maximal attainable utility is the same for every policy,
gains in $J_{j}$ equal reductions in DM~$j$'s usage risk, and maximising
$\min\{J_{1},J_{2}\}$ minimises the larger of the two usage risks.
Figure~\ref{fig:ranking-and-selection}(c) reports the gain of the average
policy in $(J_{1}+J_{2})/2$ over the specialist that is better on the
validation episodes, with a paired normal $95\%$ interval, and the gain of the
minimax policy in $\min\{J_{1},J_{2}\}$ over the best randomisation, with a
$95\%$ interval from $2{,}000$ paired bootstrap resamples stratified by test
set.

The experiment
runs on a single Apple M5 Pro CPU: the single-DM study takes about $30$ seconds
and the two-DM study about $10$ minutes.

\section{Details on Policy Learning}
\label{app:policy-learning}

Here we describe how we learn amortised design policies for the objectives
introduced in Section~\ref{sec:methodology}. That section gave a brief outline;
here we set out the Monte Carlo objective, the gradient estimators we use and
when each applies, and how the same recipe extends to the average and minimax
objectives.

We parameterise the design policy by a neural network with parameters
\(\phi\), writing \(\pi_{d}^{\phi}(x_{t} \mid D_{t-1})\) for its
distribution over the next design given the history. For a fixed downstream
rule, the objective is the usage risk~\eqref{eq:fixed-rule-risk},
\begin{equation}
    R_{\ell}(\phi) \coloneq R_{\ell}(\pi_{d}^{\phi}, \pi_{a})
    = \mathbb{E}_{p(z, D_{T}; \pi_{d}^{\phi})}
    \bigl[\ell(\pi_{a}(D_{T}), z)\bigr].
\end{equation}
We assume only that the prior \(p(z)\) and the likelihood \(p(y \mid z, x)\)
can be sampled from, and that the rule \(\pi_{a}\) can be evaluated on any
history. Simulating \(N\) independent experiments under the current policy,
\((z^{(i)}, D_{T}^{(i)}) \sim p(z, D_{T}; \pi_{d}^{\phi})\), gives the Monte
Carlo estimate
\begin{equation}
    \hat{R}_{\ell}(\phi) = \frac{1}{N} \sum_{i=1}^{N}
    \ell\bigl(\pi_{a}(D_{T}^{(i)}), z^{(i)}\bigr),
\end{equation}
and we minimise \(R_{\ell}\) by stochastic gradient descent on \(\phi\).
Since the sampling distribution itself depends on \(\phi\), the remaining
question is how to estimate \(\nabla_{\phi} R_{\ell}(\phi)\). We use the
two standard estimators~\citep{mohamed2020montecarlo}.

The first is the pathwise estimator. Suppose the designs are continuous, the
policy is deterministic or reparameterisable, the likelihood \(p(y \mid z, x)\)
is reparameterisable, and \(D_{T} \mapsto \ell(\pi_{a}(D_{T}), z)\) is
differentiable. Then, for fixed state and noise variables, the loss is a
composition of differentiable maps and hence a differentiable function of
\(\phi\), so, under the usual regularity conditions that permit exchanging
gradient and expectation~\citep{mohamed2020montecarlo}, \(\nabla_{\phi}
R_{\ell}(\phi) = \mathbb{E}[\nabla_{\phi}\, \ell(\pi_{a}(D_{T}), z)]\), and
\(\nabla_{\phi} \hat{R}_{\ell}(\phi)\) is an unbiased estimate obtained by
backpropagating through the loss, the rule, and all \(T\) rounds of the
simulator, as in deep adaptive design~\citep{foster2021deep}. This estimator
typically has lower variance than the alternative below, but requires the DM's
rule to be differentiable in the data. We use it in
Section~\ref{sec:high-dimens-image}, where the rule is a fixed classifier
applied to continuous, noisy measurements.

When instead the designs are discrete, or the loss or rule is non-differentiable
or only available as a black box, we use a stochastic policy and the
score-function, or REINFORCE, estimator~\citep{williams1992reinforce}. Since
\(\phi\) enters \(p(z, D_{T}; \pi_{d}^{\phi}) = p(z) \prod_{t=1}^{T}
\pi_{d}^{\phi}(x_{t} \mid D_{t-1})\, p(y_{t} \mid z, x_{t})\) only through the
policy factors,
\begin{equation}
    \nabla_{\phi} R_{\ell}(\phi)
    = \mathbb{E}_{p(z, D_{T}; \pi_{d}^{\phi})}
    \Bigl[\ell(\pi_{a}(D_{T}), z) \sum_{t=1}^{T}
    \nabla_{\phi} \log \pi_{d}^{\phi}(x_{t} \mid D_{t-1})\Bigr],
\end{equation}
which is estimated from the same \(N\) simulated trajectories. The rule and the
loss now only need to be evaluated, never differentiated. The price is variance,
which we reduce in the usual ways: replacing the loss in the \(t\)-th term of
the sum by \(\ell(\pi_{a}(D_{T}), z) - b(D_{t-1})\) for a baseline \(b\) that
depends only on the history $D_{t-1}$, which leaves the expectation unchanged; standardising
the resulting advantages within each batch; and adding an entropy bonus to
prevent premature collapse to a deterministic policy. This was sufficient in
Section~\ref{sec:clinical-trials}, where the regulator's test is treated as
a black box~(Appendix~\ref{app:clinical-trials}).

More generally, minimising \(R_{\ell}\) is a finite-horizon reinforcement
learning problem: the state is the history \(D_{t-1}\), the action is the design
\(x_{t}\), transitions are given by simulating the experiment, and the only
reward is the terminal \(-\ell(\pi_{a}(D_{T}), z)\), so any policy-gradient
method applies. In Section~\ref{sec:destructive-lot-inspection} we use proximal
policy optimisation~\citep[PPO,][]{schulman2017proximal}, which adds two
stabilising ingredients to REINFORCE: a learned critic \(V_{w}(D_{t-1})\), which
estimates the expected terminal reward from a state and serves as the baseline,
and a clipped surrogate objective, which lets each batch of trajectories be
reused for several minibatch updates while keeping the policy close to the one
that generated the data. Hyperparameters are given in
Appendix~\ref{app:destructive-lot-inspection}. Learned policies are not always
the best option, however: in Section~\ref{sec:ranking-and-selection}, neither
REINFORCE nor PPO matched the usage-aware knowledge-gradient
policy~(Appendix~\ref{app:ranking-and-selection}).

\subsection{Uncertain Downstream Rules}

While we do not learn neural policies for the average or minimax objectives in
this paper, both fit the same recipe. The average
objective~\eqref{eq:bayes-average-objective} differs
from~\eqref{eq:fixed-rule-risk} only in that each simulated trajectory draws the
pair \((z, \psi)\) jointly from \(p(z, \psi)\), rather than \(z\) from \(p(z)\),
and reads off the loss with \(\pi_{a}^{\psi}\). Because \(\psi\) is not observed
during data collection, the policy factors of the trajectory distribution are
unchanged, and both gradient estimators above carry over verbatim, the pathwise
one provided \(\pi_{a}^{\psi}\) is differentiable for almost every \(\psi\).

The minimax objective~\eqref{eq:minimax-objective} replaces the expectation over
\(\psi\) by a supremum, which is not an expectation over trajectories. Writing
\(R_{\psi}(\phi)\) for the inner expectation, when \(\Psi\) is finite or
discretised we can relax the supremum to the smooth upper bound \(\tau^{-1} \log
\sum_{\psi} \exp(\tau R_{\psi}(\phi))\), whose gradient is \(\sum_{\psi}
w_{\psi} \nabla_{\phi} R_{\psi}(\phi)\) with \(w = \operatorname{softmax}(\tau
R(\phi))\). Each step is then an average-objective step whose weights
concentrate on the rules the policy currently serves worst, and \(\tau \to
\infty\) recovers the supremum. In practice the weights are computed from Monte
Carlo estimates of the \(R_{\psi}(\phi)\), through which they enter nonlinearly,
so the resulting gradient estimate is biased, though consistent as \(N \to
\infty\). For continuous \(\Psi\), one instead alternates descent steps on
\(\phi\) with ascent steps for an adversary over \(\psi\). More generally,
fixing \(\phi\), each rule \(\psi\) yields a usage risk \(R_{\psi}(\phi)\).
The average objective minimises a weighted sum of these risks and the minimax
objective their maximum, which are the two standard scalarisations in
multi-objective reinforcement learning, so methods from that literature apply
directly~\citep{roijers2013survey, hayes2022practical}.

\section{Limitations and Broader Impact}
\label{app:broader-impact}

\subsection{Limitations}
\label{app:limitations}

\paragraph{Access to the downstream rule.}
Usage-aware design requires access to the downstream rule or to a collection
of plausible rules that can be evaluated on simulated data. When the rule is
uncertain, the experimenter must specify a prior over these rules or a set
against which to optimise robustly~(Section~\ref{sec:methodology}). The resulting
design is only as good as this description. A design tailored to one rule can
perform poorly under another, as the transferred policies in
Section~\ref{sec:high-dimens-image} show, and the minimax
objective~\eqref{eq:minimax-objective} protects only against rules in the chosen
set~\(\Psi\).

\paragraph{Scope of the theory.}
Proposition~\ref{prop:blackwell-safe-rule} and
Corollary~\ref{cor:epu-representation} assume a finite state space
\(\mathcal{Z}\), a prior with full support, and a rule that acts on the data
only through the experimenter's posterior. They do not cover rules that use
information the posterior discards, such as a classifier applied directly to
raw measurements, although the objectives of Section~\ref{sec:methodology} and
the decomposition of Section~\ref{sec:incompatibility} hold for any rule.
Extending the characterisation to infinite \(\mathcal{Z}\) is not
straightforward, since there concavity no longer implies Jensen's
inequality~\citep{bect2026uncertainty}.

\paragraph{Optimisation.}
Downstream rules are often discontinuous, as with tests and thresholds, so
pathwise gradients are frequently unavailable and policies must be trained with
higher-variance score-function estimators~(Appendix~\ref{app:policy-learning}).
Training also requires many simulated episodes, which can be expensive when
the rule itself is costly to evaluate. In Section~\ref{sec:ranking-and-selection},
neural policies trained directly on the usage risk did not match an
adapted knowledge-gradient policy.

\subsection{Broader Impact}

Our framework puts no restriction on the downstream rule. In many cases, the DM
has preferences of their own, encoded in their own prior \(q(z)\) and loss
\(\ell_{\mathrm{DM}}\), as in Section~\ref{sec:destructive-lot-inspection},
while the experimenter's loss may reward actions the DM would regret. Can the
experimenter then persuade the DM to act against their own interests, and if
so, by how much?

This is the question studied in Bayesian
persuasion~\citep{kamenica2011bayesian}, and in our framework the answer hinges
on one feature. The experimenter chooses the designs, but the outcomes come
from a likelihood \(p(y \mid z, x)\) that they cannot alter, and the DM sees all
of the data, with nothing withheld or cherry-picked. Suppose the DM shares this
likelihood.\footnote{The bound in~\eqref{eq:dm-no-harm} also holds with
\(p(y \mid z, x)\) replaced by any likelihood the DM assumes, but then only
under the DM's own model of the data.} Since each design depends only on earlier data, the policy terms in
\begin{equation}
    \label{eq:history-likelihood}
    p(D_{T} \mid z; \pi_{d})
    = \prod_{t=1}^{T} \pi_{d}(x_{t} \mid D_{t-1})\, p(y_{t} \mid z, x_{t})
\end{equation}
do not involve \(z\) and cancel in Bayes' rule. The same holds for a stopping
rule that depends only on the data, like the manufacturer's in
Section~\ref{sec:destructive-lot-inspection}, since the probability of stopping
at each round then does not involve \(z\) either and also cancels. The DM's posterior
\(q_{z}(\cdot \mid D_{T})\) is therefore correct whichever policy produced the
data, and the DM need not even know the policy. The DM's posterior averages
back to their prior, so a DM who takes the Bayes act
\(a_{\mathrm{DM}}^{*}(D_{T})\) under their own beliefs cannot, in expectation,
be harmed by the data~\citep{kamenica2011bayesian}:
\begin{equation}
    \label{eq:dm-no-harm}
    \mathbb{E}_{q(z, D_{T}; \pi_{d})}\bigl[
    \ell_{\mathrm{DM}}(a_{\mathrm{DM}}^{*}(D_{T}), z)\bigr]
    =
    \mathbb{E}_{q(D_{T}; \pi_{d})}\Bigl[
    \mathbb{H}_{\ell_{\mathrm{DM}}}\bigl[q_{z}(\cdot \mid D_{T})\bigr]\Bigr]
    \leq
    \mathbb{H}_{\ell_{\mathrm{DM}}}[q],
\end{equation}
where \(q(z, D_{T}; \pi_{d}) \coloneq q(z)\, p(D_{T} \mid z; \pi_{d})\) and the
inequality is Jensen's, as \(\mathbb{H}_{\ell_{\mathrm{DM}}}\) is concave.
Whatever the experimenter wants, the DM expects to do at least as well as
without the experiment. Withholding data breaks this. If the experimenter
reports only part of what they collected, the probability of the report depends
on how that part was chosen, so a DM who takes it at face value has the wrong
posterior and loses the guarantee. Bayesian persuasion avoids this by assuming
that the receiver knows exactly how the sender's signal is generated. Our
framework needs no such assumption, because the only choices the experimenter
makes are the designs, and the DM sees them.

The guarantee still leaves room for influence in two ways. First, it only
compares with having no data: among informative experiments, the experimenter
may pick those that suit them rather than the DM, which
\citet{adusumilli2026designing} suggest a regulator could curb with a floor on
the DM's expected welfare. Second, it requires the DM to take the Bayes act.
Under other rules, more data can hurt the DM too: applying
Proposition~\ref{prop:blackwell-safe-rule} with the DM's loss, no experiment
can harm the DM, whatever their prior, if and only if \(r_{\pi_{a}}\) is concave.
Persuading a DM into harm thus requires a rule that departs from the Bayes act,
as with the non-Bayesian receivers of~\citet{declippel2022nonbayesian}, and
Proposition~\ref{prop:blackwell-safe-rule} tells us which rules are vulnerable.

\end{document}